\documentclass[11pt]{article}

\usepackage{microtype} 
\usepackage{booktabs}  
\usepackage{url}  
\usepackage[subpreambles=true]{standalone}
\usepackage{import}
\usepackage{tikz}
\usepackage{amsmath}
\usepackage{amsthm}

\usepackage{tikz}
\newcommand*\circled[1]{\tikz[baseline=(char.base)]{
            \node[shape=circle,draw,inner sep=1pt] (char) {#1};}}

\newcommand{\structure}[1]{\todo[inline, color=red!40]{#1}}  
\usepackage[final]{automl}
\newcommand{\methodname}{TACTICL\xspace}
\newcommand{\method}{adaptation\xspace}

\newcommand{\adapter}{adapter\xspace}
\newcommand{\adapters}{adapters\xspace}
\newcommand{\Adapter}{Adapter\xspace}
\newcommand{\tabarena}{47}
\newcommand{\tabarenac}{34}
\newcommand{\tabarenar}{13}
\newcommand{\tabarenaname}{TabArena}

\newcommand{\tabpfn}{TabPFN}
\newcommand{\tabpfnone}{TabPFNv1}
\newcommand{\pfntwohalf}{TabPFNv2.5}
\newcommand{\pfntwo}{TabPFNv2}
\newcommand{\pfnthree}{TabPFNv3}
\newcommand{\tabicl}{TabICL}
\newcommand{\tabicltwo}{TabICL2}
\newcommand{\codeurl}{\url{https://github.com/Hebog/tfm_compression}}

\usepackage[capitalize,noabbrev]{cleveref}
\usepackage{placeins}
\usepackage{makecell}
\usepackage{wrapfig}

\usepackage{natbib}
\usepackage[disable]{todonotes}

\title{TACTICL: Task-Aware Compression of Tabular ICL Models}

\author[1,2]{\nameemail{Mykhailo Koshil}{mykhailo.koshil@tu-dortmund.de}}
\author[1,2]{\nameemail{Matthias Feurer}{matthias.feurer@tu-dortmund.de}}
\author[1,2]{\nameemail{Katharina Eggensperger}{katharina.eggensperger@tu-dortmund.de}}

\affil[1]{TU Dortmund University, Germany}
\affil[2]{Lamarr Institute for Machine Learning and Artificial Intelligence, Germany}

\hypersetup{%
  pdfauthor={}, 
  pdftitle={},
  pdfsubject={},
  pdfkeywords={}
}

\begin{document}

\maketitle

\begin{abstract}
The strong performance of foundation models for tabular tasks comes at substantial inference costs. Distilling models into task-specific architectures reduces model size and computational demands but also sacrifices in-context adaptability. Here we introduce \methodname{}, an automated task-aware compression framework for tabular in-context learning models that jointly prunes transformer layers and replaces them with lightweight adapters trained on downstream tasks, thus blending in-context with in-weight learning. We study \methodname{} on \tabarena{} benchmark datasets and show that we can substitute up to $85$\% of layers without substantial performance drop on a given downstream task. We further show that \methodname{} maintains robustness to data shifts, leaving its in-context ability intact. Overall, \methodname{} provides a robust framework for exploiting the depth-wise redundancy of tabular foundation models by combining task-specific adaptation and structured compression. We provide the code at: \codeurl{}.
\end{abstract}


\section{Introduction}

\noindent Tabular foundation models (TFMs) have advanced rapidly, with successive generations achieving state-of-the-art performance while also growing in size and inference cost. 
\tabpfnone{}~\citep{tabpfn} comprises $\sim$12M parameters and was limited to tables with 1,000 samples and 100 features. 
\pfntwohalf~\citep{tabpfntwohalf} consists of 10M weights and "was built for datasets with up to 50,000 data points and 2,000 features", and \tabicltwo{}~\citep{tabicl2} consists of 27M weights and "generalizes effectively to million-scale datasets". While applicability increases,  
inference costs are still half an order of magnitude larger than lightweight tree-based baselines~\citep{tabicl2}.\footnote{~\cite{gangwani2025light} reported inference costs of \pfntwo{} to be $>2\,000\times$ slower than XGBoost (and up to $11\,000\times$ for \tabicl{}), while requiring up to 9\,GB of GPU VRAM compared to $<150$\,MB of RAM for tuned tree ensembles.} 

To reduce inference costs, distillation into task-specific models has emerged as a resource-efficient deployment strategy for ICL-based tabular models. 
\begin{wrapfigure}{r}{0.5\textwidth}
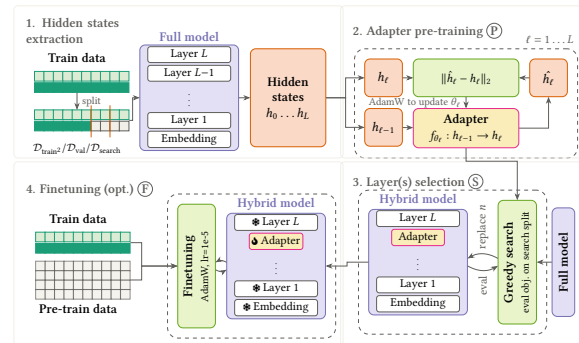
 
    \centering
    \resizebox{0.5\textwidth}{!}{
        \includestandalone{figs/algo}
        }
    \caption{Overview of \methodname{}, which jointly compresses and adapts \tabpfn{} models.}
    \label{fig:overview}
\end{wrapfigure}
All \pfntwo{}~\citep{hollmann-nature25a}, \pfntwohalf{}~\citep{tabpfntwohalf}, and \pfnthree{}~\citep{tabpfn3} introduce dedicated distillation engines that produce, e.g., a compact multi-layer perceptron (MLP) for a given dataset with orders-of-magnitude lower latency. While effective, this approach explicitly converts the model to the in-weight-learning (IWL) regime, encoding the training distribution directly into weights. This means the model must be re-distilled from scratch whenever the context changes, forfeiting an attractive property of in-context learning (ICL): the ability to adapt to new data at inference time without retraining.
Beyond zero-shot usage, when predictive performance is a primary objective, adapting TFMs to downstream tasks has become a common strategy, as demonstrated by RealTabPFN~\citep{gargreal} and others~\citep{rubachev2025finetuning,kolberg2026tabpfnwidecontinuedpretrainingextreme}, in which continued pre-training on in-domain data yields consistent accuracy gains. 

At the same time, researchers started studying the layer impact of TFMs and found that not all layers contribute equally to task performance~\citep{balef2025towards,balef2025isone}, which is in line with findings for language ICL models \citep{lad2025remarkable,men2025shortgpt,gromov2025unreasonable}. Concretely, they found that performance is not substantially affected by removing, repeating, or swapping a single layer. This hints at depth-wise redundancy, and pruning layers offers an opportunity to substantially lower inference costs without sacrificing accuracy. 

Based on this motivation, we study: \textit{Can we compress TFMs without collapsing them into pure in-weight learners?} Specifically, we study whether we can simultaneously prune and adapt a TFM for a downstream task, achieving the benefits of both compression and fine-tuning within a single framework and without sacrificing ICL capability. This is non-trivial for several reasons: Firstly, prior work shows that the transition from in-context to in-weights learning is sharp and often irreversible~\citep{singh2023transient}, and fine-tuning a model on a fixed dataset can actively suppress its ICL ability due to low data diversity~\citep{nguyen2025differential,bethune2025scaling}. Secondly, we found that model manipulation is highly task-specific; thus, the optimal pruning strategy depends on the dataset at hand and requires automation to be applied robustly.



We propose \methodname{} for \textit{task-aware compression of tabular in-context models}, which identifies layers to prune and replaces them with lightweight adapter modules, trained on the downstream task (see overview in Figure~\ref{fig:overview}). In contrast to distillation, our method retains the full flexibility of the original model: it preserves in-context learning capabilities, remains robust to perturbations, and can be applied to unseen tasks without retraining from scratch. \methodname{} automatically optimizes a compression configuration, even for large models where an exhaustive search would be intractable. Our contributions are as follows:
\setlist{nolistsep}
\begin{itemize}[noitemsep]
    \itemsep0em 
    \item A \textit{unified framework} for simultaneous structured pruning and domain \method{} for state-of-the-art tabular ICL models.
    \item The \textit{first automated pipeline} for explicitly combining in-context learning and dataset-specific memorization to decrease computational complexity while maintaining or even improving predictive performance.
    \item A \textit{quantitative evaluation} on a subset of \tabarena{} \tabarenaname{} datasets~\citep{erickson2026tabarena}, demonstrating that \methodname{} can retain the performance of \pfntwohalf{} while reducing the compute cost almost linearly with the number of dropped layers.
\end{itemize}

\section{Problem Setup: Task-aware Model Compression}
\structure{
   Preliminaries:
model compression and ft: problem definition
related work
Parameter-efficient adaptation \\
Layer redundancy and structured compression: investigate removing layers according to the embedding similarity: unreasonable ineffectiveness of deeper layers)\citep{gromov2025unreasonable}; short gpt \citep{men2025shortgpt}; STREAMLINING REDUNDANT LAYERS TO COMPRESS LARGE LANGUAGE MODELS \citep{chen2025streamlining}
Fine-tuning dynamics and forgetting
Layer sensitivity analysis
}

\noindent We start by formally defining our problem setup. Given a pretrained model $\mathcal{M}$ with $L$ layers, e.g., transformer blocks, and a supervised learning downstream task, represented by labelled training $\mathcal{D}_{train}$, hold-out (search) $\mathcal{D}_{search}$ and unlabelled test data $\mathcal{D}_{test}$ sets.  
For $\mathcal{M}$ with $L$ layers, we define a compression configuration $c \in \{0,1\}^L$ as a binary vector indicating which layers are retained ($c_i = 1$) or removed ($c_i = 0$). Given a target compression ratio $n/L$ when dropping $n$ out of $L$ layers, the set of all valid configurations: 
\begin{equation}  
    C_n = \left\{ c \in \{0,1\}^L \;\middle|\; \sum_{i=1}^{L} c_i = L - n \right\}. 
    \label{eq:conf}
\end{equation}
We refer to $\mathcal{M}_c$ as a compressed model using configuration $c$ and measure generalization performance as $f(c) = \mathcal{L}\left(y_{\it search}, (\mathcal{M}_c(x_{\it search} | D_{\it train}) \right)$ with $\mathcal{L}$ being a metric suitable for the task type, e.g., AUC for classification and RMSE for regression. 
The goal is then to identify $c^*$ which maximizes this score (or minimizes in case of a loss) for a fixed compression ratio $n/L$ 
for a given  $\mathcal{D}_{\it train}$:
\begin{equation}
    c^* \in \operatorname*{argmax}_{c \,\in\, C_n} f(c).
\end{equation}

Alternatively, instead of removing layers, we may substitute a layer with a lightweight \adapter{}, e.g., a shallow MLP, trained on a subset of $\mathcal{D}_{train}$ to approximate the original layer's input-output hidden-state mapping to \textit{heal} the damage introduced by removing a layer (leaving the problem formalization \eqref{eq:conf} equivalent for substitution).  We refer to this automated compression process as \textit{\method{}}, as it tailors the model $\mathcal{M}$ to a task by optimizing the compression configuration per task and replacing task-agnostic layers with adapter modules, trained on downstream data.

\subsection{Background and Related Work}
Our approach sits at the intersection of three lines of work: parameter-efficient adaptation, 
structured model compression, fine-tuning dynamics, and prior work in the tabular domain, which we will briefly discuss in the following.

\textbf{Parameter-efficient adaptation.}
\citet{houlsby2019parameter} introduced adapter modules as a parameter-efficient alternative to full fine-tuning, inserting small bottleneck networks between the frozen layers of a pretrained NLP transformer. This creates a single shared backbone specialized to new tasks by training only a few additional parameters per task. Our use of lightweight substitute modules to replace dropped layers is conceptually related, but differs in purpose: rather than adapting a frozen model to new tasks, we use adapters to \emph{restore} the computation lost by layer removal while simultaneously enabling domain adaptation. Complementary to this, \citet{guo2021parameter} proposed diff pruning, which frames fine-tuning as learning a sparse task-specific \emph{diff vector} on top of the frozen model. The diff vector is pruned via a differentiable $L_0$-norm approximation, yielding models that match full fine-tuning performance while modifying fewer than $0.5\%$ of the frozen weights per task. Diff pruning shares our interest in minimizing the parameter cost of adaptation, but it operates by selectively updating weights rather than by replacing layers.

\textbf{Layer redundancy and structured compression.}
A growing body of work demonstrates that transformer layers are substantially more redundant than one might expect, motivating layer-level pruning as a principled compression strategy. This redundancy is frequently explained by the circuit hypothesis \citep{elhage2021mathematical}, which holds that task-specific behaviour is implemented by sparse, localized subnetworks of attention heads and MLPs rather than the full depth of the model, so layers outside these task-relevant circuits can be pruned with little effect on performance.
\citet{muralidharan2024compact} proposed a complementary multi-axis compression strategy that jointly prunes model components (depth, width, attention heads, and MLP units) and then retrains the pruned model via knowledge distillation on less than 3\% of the original pretraining data. This yields models that are 2-4$\times$ smaller while remaining competitive.
\citet{li2026distilling} introduced a compression method that reduces inference cost by selectively distilling softmax attention layers into linear attention. They showed that per-layer KL divergence between the teacher and a linear-attention student provides an effective score for determining which layers can be replaced, enabling significant compression while recovering teacher-level performance through distillation-based fine-tuning. Both approaches share a common structure with our method: identify less important components via a principled score, remove or replace them, and recover performance through lightweight fine-tuning. 
\citet{gromov2025unreasonable} showed that for popular open-weight LLMs, one can remove up to half of the deepest layers, identified by computing the angular distance between adjacent layer representations, with minimal degradation on question-answering benchmarks, provided the damage is subsequently ``healed'' via a small amount of parameter-efficient fine-tuning. This work directly motivates our analysis of proxy metrics for layer sensitivity and our usage of adapters. 
Similarly, \citet{men2025shortgpt} introduced Block Influence (BI), a training-free importance metric that requires only a single forward pass over a small calibration set: it measures how much each layer transforms its input hidden state, meaning layers with low BI scores are close to identity transformation; layers with low BI scores are dropped, yielding competitive compression without gradient computation or model retraining. 
\citet{chen2025streamlining} extended this direction by replacing entire redundant blocks from LLMs with a lightweight adapter, further demonstrating that structured, coarse-grained pruning at the layer level is a practical and effective compression strategy. In a similar fashion, \citet{shopkhoev2025replaceme,cannistraci_tmlr26a} maintain model performance with substantial compression by utilizing linear adapters.

In contrast to these works, which mostly target general-purpose LLMs and recover performance via healing, our setting introduces two additional challenges: First, tabular foundation models perform \emph{in-context learning} over tabular data, and the sensitivity of layers to removal differs substantially from autoregressive LLMs~\citep{balef2025isone}. 
Second, we aim for simultaneous compression \emph{and} domain adaptation, rather than recovery to the original distribution.

\textbf{Fine-tuning dynamics.}
When a pretrained model is fine-tuned on a target domain, it risks two failure modes: overfitting to scarce target data and forgetting generic capabilities acquired during pretraining. \citet{bethune2025scaling} derived scaling laws that jointly quantify these two phenomena as a function of model scale, available fine-tuning data, and the fraction of pretraining data re-injected into the fine-tuning mixture. A key finding is that injecting as little as $1\%$ of pretraining data is sufficient to prevent catastrophic forgetting. This result is directly relevant to our setting: when fine-tuning the pruned and adapter-augmented model for domain adaptation, we risk degrading the ICL capabilities of the retained transformer layers. We use this insight to design our fine-tuning protocol
and motivate our out-of-distribution (OOD) evaluation, which explicitly tests whether the compressed model retains the original model's ability to perform ICL.

\textbf{Tabular domain.} \citet{kuken2025early}  make use of an early-exit mechanism for TFMs, attaching lightweight decoders to each transformer layer that allow inference to terminate early based on prediction entropy, with no task-specific fine-tuning. This yields inference speed-ups of up to 1.3 to 2.2 times with small accuracy loss. Similarly, \citet{liu2026tabswift} make use of an early exit. However, such an approach is a dynamic inference technique rather than model compression: the full backbone is retained, and no parameters are pruned, quantized, or distilled. 

\todo[inline]{Describe the difference to model distillation}
\todo[inline]{Cite LEMONADE (Thomas' paper on NAS with Lamarckian evolution)}

\section{Methodology for \methodname{}}

To simultaneously prune and adapt, we design a framework consisting of three components: \circled{P} training lightweight adapters, \circled{S} searching for the optimal pruning configuration, and \circled{F} fine-tuning the whole model. If adapters are not used, only the second and third components are needed. In this section, we will describe the challenges and our approach for each component and begin with the core algorithm of our work, \circled{S}.

\textbf{\circled{S} Layer Selection.} Prior research found that transformer layers in tabular models contribute unequally to in-context prediction, with some being close to redundant for some tasks~\citep{balef2025isone}. Thus, such layers may be removed or approximated by a simpler function. Since we aim to find an optimal compression configuration, we need to efficiently search over $2^L$ pruning configurations for each task.
As a solution, we propose a greedy search over possible layer-replacement configurations, guided by AUC/RMSE and prediction stability evaluated on a held-out search set. This means we iteratively drop the layer with the least negative impact on empirical generalization performance. 
We run a proof-of-concept experiment on \pfntwo{} to study the effectiveness of our selection method. We find that the pruning scheme is dataset-dependent
(see Section~\ref{sec:poc}), and show that our greedy strategy recovers a near-oracle configuration (see Section~\ref{sec:search}).

\textbf{\circled{P} \Adapter\ Pre-Training.} Instead of dropping, we consider replacing layers with \adapter models: For each dropped layer, we train a small, lightweight model to approximate the functionality of the layer it replaces via supervised learning on the original in- and output embeddings. 
In addition to the task-aware layer configuration, this further introduces another task-dependent component that further adapts the model to the downstream task.
\label{sec:adapter_training}

Using adapters introduces two challenges. First,
we must consider its architecture, hyperparameters, and training protocol. Secondly, the optimal layers to drop need not be the optimal layers to replace with \adapters (as we show in Section~\ref{sec:search}).\footnote{For example, a layer that implements a simple yet crucial function can be replaced by an adapter, whereas removing it would severely degrade performance. } As replacing self-attention by $1 \times 1$ convolution reduces expressivity by eliminating inter-token interactions, in-context learning can be negatively impacted. Therefore, this loss has to be offset by dataset-specific knowledge learned during adapter training, underscoring the need for task-aware compression.
As a practical solution, we use a lightweight adapter architecture:
a two-layer MLP with GELU activations and a skip connection, selected as a result of a small architecture search; more details in Appendix~\ref{app:adapter-nas}.

We pretrain all adapters once, independently and in parallel, before the search begins, rather than retraining an adapter at every search step (Figure~\ref{fig:overview}); although sequential per-step training would in principle let each adapter correct for errors introduced by previously substituted layers, upfront pretraining substantially cuts search cost by incurring the expensive adapter training once per layer rather than repeatedly at each iteration. This choice follows from a locality assumption underlying our greedy search, i.e., that the decision to keep or delete/substitute a given layer does not depend on the decision for other layers.

\textbf{\circled{F} Fine-Tuning.} As a final step, we optionally fine-tune the compressed model to align layers and adapters and recover residual performance lost during compression. This is straightforward and similar to prior work to prevent overfitting~\citep{bethune2025scaling}: we run a few epochs of AdamW to update the weights of the adapter models (with all other weights frozen), using a mixture of pretraining data and data from the downstream task. Freezing the transformer layers is a deliberate design choice that confines learning to the in-weights pathway, leaving the in-context inference circuitry of the retained layers intact, thereby preserving the ICL capabilities. This is consistent with the theoretical view that the ICL and IWL sub-circuits are largely independent~\citep{nguyen2025differential}, such that updating only adapter weights steers the in-weights pathway without disrupting in-context processing, as we show in Section~\ref{sec:not-mlp} and Section~\ref{sec:finetuning}. 

\section{Experiments}\label{sec:experiments}

We now turn to the empirical evaluation of our \methodname{}. We start with a straightforward proof-of-concept experiment to validate and study our method under controlled conditions in a feasible setup, and then assess performance on state-of-the-art models.
Specifically, we organize this section around five research questions:
(Section~\ref{sec:poc})~\textbf{Can less be more?} explores whether layer removal/substitution can exploit redundancy to preserve performance;
(Section~\ref{sec:search})~\textbf{Is greedy good enough?} studies whether our local search heuristic can recover near-oracle compression configurations efficiently; 
(Section~\ref{sec:not-mlp})~\textbf{Can it still learn in-context?} assesses whether the resulting model is robust against reasonable domain shifts and perturbations; 
(Section~\ref{sec:finetuning})~\textbf{Do we need fine-tuning?} evaluates whether end-to-end fine-tuning recovers performance loss due to compression; and, finally, (Section~\ref{sec:speedup})~\textbf{Does \methodname{} speed up predictions?} quantifies inference speed-ups. \methodname{} can be applied to any supervised learning task; in this section, we focus on discussing classification results and report regression results in the Appendix.
%
%

\vspace{0.1cm}
\noindent\textbf{Datasets and Evaluation Protocol.} We measure in-distribution performance using \tabarenac{} classification and \tabarenar{} regression datasets from the \tabarenaname{} benchmark~\citep{erickson2026tabarena}. We chose a consistent subset of tasks for all experiments, which consists of all tasks natively supported by \pfntwohalf{} after splitting into train and test subsets. We use a single fold for \pfntwo{}, three folds for \pfntwohalf{}, and report metrics aggregated across datasets. We assess out-of-distribution generalization in Section~\ref{sec:not-mlp} by applying three controlled perturbations to the in-distribution test sets: permuting the feature order, injecting uninformative features, and flipping the labels. To assess speed-ups, we average inference time across 6 repeated inferences preceded by one not-measured warm-up inference. We provide additional details in Appendix~\ref{app:datasets}.

\noindent\textbf{Hardware and Resource Consumption.} We ran all experiments on an NVIDIA B300 in approximately 6 days of GPU time. Additionally, inference time measurements and development was done on NVIDIA DGX Spark with 3 days of GPU time in total. 

\subsection{Can Less Be More? Proof-of-Concept on \Adapter{} Substitution}
\label{sec:poc}

\structure{goal: show that replacing a layer with an \adapter{} maintains model performance; the exhaustive-search experiment and its results.}

We build on prior work showing that layers contribute unequally to in-context prediction~\citep{balef2025isone} and, here, establish how much the model can be compressed while preserving performance, using \pfntwo{} as a small, yet capable model. For this, we exhaustively evaluate all possible compression configurations, sweeping the number of dropped or substituted layers from $0$ to $L=12$. We train $L=12$ adapters offline beforehand and exhaustively evaluate all $4096$ configurations. Specifically, for each configuration size $n$ (i.e., $n$ pruned or substituted layers), we enumerate all $\binom{12}{n}$ configurations, measure the maximum change in AUC/RMSE relative to the full model, and report the median, maximum, and minimum of this metric across all datasets. This allows capturing the best-case to worst-case improvement achievable at each level of compression.

\begin{figure}[tb]
    \centering
    \begin{subfigure}{0.4\linewidth}
        \includegraphics[width=\linewidth]{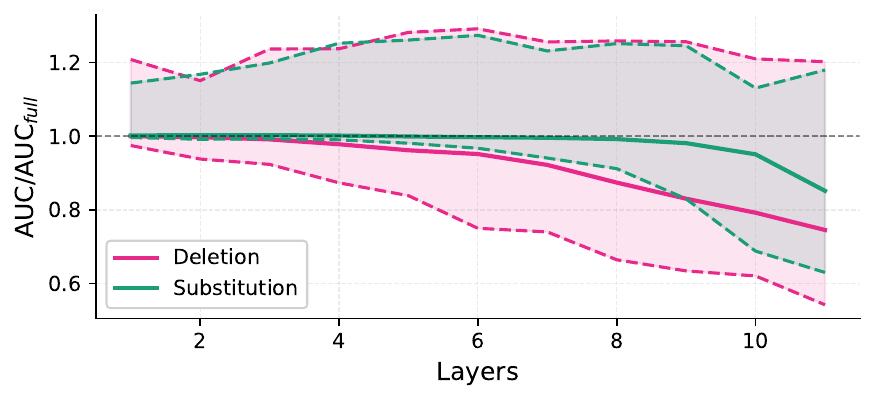}
        \caption{Classification}
    \end{subfigure}
    \begin{subfigure}{0.4\linewidth}
        \includegraphics[width=\linewidth]{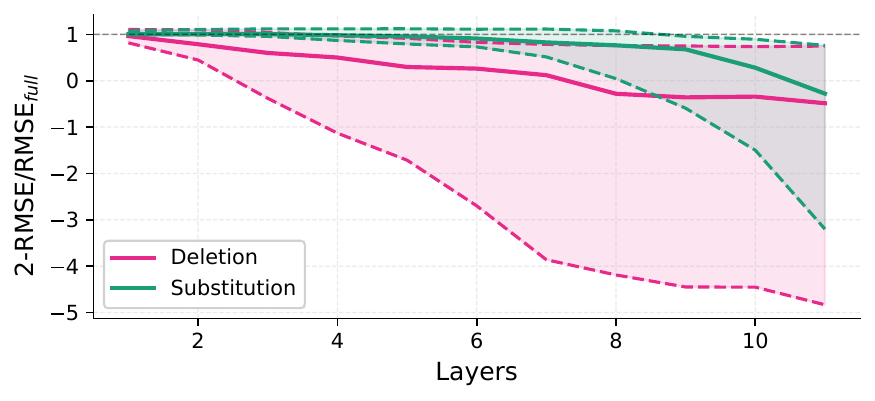}
        \caption{Regression}
    \end{subfigure}
    \caption{Minimal, maximal, and Median of maximal relative performance change over \tabarenac{} classification and \tabarenar{} regression datasets from \tabarenaname{}
    as a function of the number of layers dropped/substituted with adapters compared to the full model (black dotted line). This plot can be interpreted as worst/best/median performance of an oracle search algorithm.}
    \label{fig:max-improvement}
\end{figure}

Figure~\ref{fig:max-improvement} reports results for dropping/substituting layers for classification and regression tasks. Most notably, dropping up to 3 layers consistently yields a non-negative performance change for half of the datasets, confirming redundancy. The substitution results are even more positive, showing that we can substitute up to 9 layers with non-negative performance changes in some settings. Unsurprisingly, as more layers are pruned, performance decreases. Overall, this shows that optimally dropping and substituting layers bears great potential to reduce model size and inference cost with minimal or no impact on final performance.

Furthermore, to motivate the need for the search for the optimal compression configuration, we calculate Shapley values per layer and show them in Figure~\ref{fig:shapley} in Appendix~\ref{app:shapley} to study layer importance. We observe that while some layers seem safe to remove or substitute across datasets, results vary greatly between removing and substituting layers as well as across datasets, serving as a strong motivation for task-aware compression.

\subsection{Is Greedy Good Enough? Efficient Compression Configuration Search}
\label{sec:search}
\structure{Design of the greedy search method and proxy alternatives; evaluation against the oracle from exhaustive search; results on TabPFN~v2.}

\noindent Next, we study how to find a dataset-specific pruning configuration in practice. Specifically, we propose a greedy search guided by a performance proxy. 
We demonstrate that it finds near-optimal configurations in our small-scale setting as well as on state-of-the-art models.

\textbf{Setup.} We split the training data $\mathcal{D}_{\rm train}$ into three disjoint sets with an 8:1:1 ratio: $\mathcal{D}_{\rm train} = \mathcal{D}_{\rm train^2} \cup \mathcal{D}_{\rm val} \cup \mathcal{D}_{\rm search}$ (see Figure~\ref{fig:overview}). $\mathcal{D}_{\rm train^2}$ and $\mathcal{D}_{\rm val}$ are used as context and query data to extract hidden states for \adapter{} training, while $\mathcal{D}_{search}$ is used to evaluate candidate configurations. At each step, configurations are scored using either lightweight structural proxies that require only a forward pass on $\mathcal{D}_{\rm val}$, or the AUC/stability metric computed directly on  $\mathcal{D}_{\rm search}$. 
At test time, all three splits are merged into a single context for inference.

\begin{figure}[t]
    \centering
        \begin{subfigure}{0.4\linewidth}
        \includegraphics[width=\linewidth]{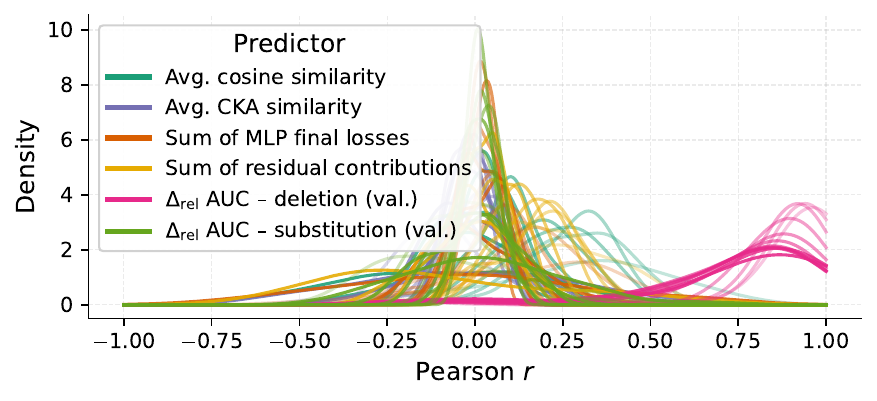}
        \caption{Deletion}
    \end{subfigure}
    \begin{subfigure}{0.4\linewidth}
        \includegraphics[width=\linewidth]{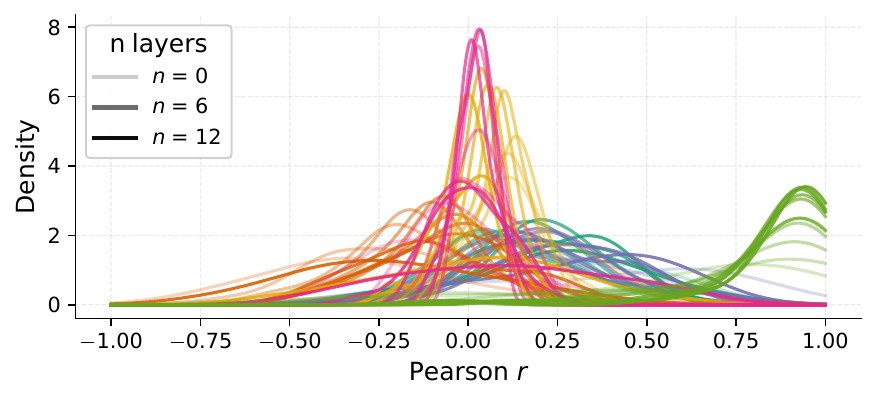}
        \caption{Substitution}
    \end{subfigure}
      \caption{Correlation between proxy scores and AUC for deleting (left) and substituting (right) $n$ layers of \pfntwo{} (across all configurations in $C_n$; see Eq.\eqref{eq:conf}), calculated from exhaustive evaluation in Section \ref{sec:poc}. None of the proxies reliably predict the performance impact, consistent with findings in \citet{sajjad2023effect}. Note that the best layers for dropping are not indicative of the best layers for substitution and vice versa. For regression results, see Figure~\ref{fig:proxies-vs-RMSE} in Appendix~\ref{app:x-plots}.}
    \label{fig:proxies-vs-auc}
\end{figure}

\textbf{Performance Proxies.} We report the correlation between several proxy metrics evaluated on $\mathcal{D}_{\rm search}$ (as described in Appendix~\ref{app:proxies}) and test performance in Figure~\ref{fig:proxies-vs-auc}. We observe that none of the structural proxies reliably predict the performance impact of layer removal or substitution. The failure is most pronounced for deletion: analysis of the data showed that layer~0, which is among the most critical layers (as established by a Shapley analysis in Figure~\ref{fig:shapley}
), is consistently assigned high proxy scores, indicating low sensitivity---precisely the opposite of the ground truth. The only exception is AUC as measured on the search set, which retains meaningful predictive performance and motivates its use as the guiding criterion in the greedy search. In search of a better proxy, we adapt and evaluate the stability metric by \citet{chen2025streamlining}, but find that it does not perform well for deletion and provide details in Appendix~\ref{app:proxies}.

\begin{figure}[b]
    \centering
        \centering
        \begin{subfigure}{0.48\linewidth}
        \includegraphics[width=\linewidth]{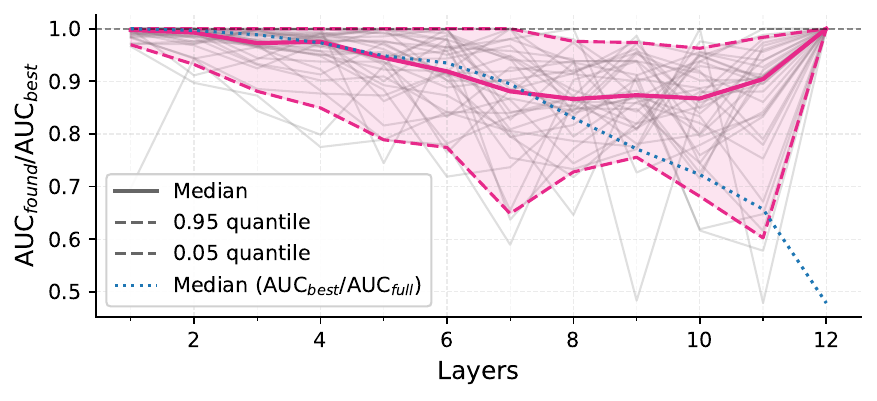}
        \caption{Deletion}
        \label{fig:exa-online-cls-del}
    \end{subfigure}
    \begin{subfigure}{0.48\linewidth}
        \includegraphics[width=\linewidth]{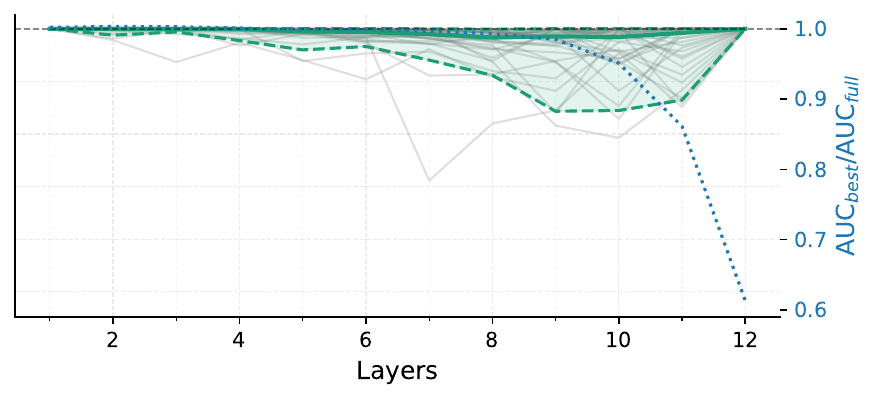}
        \caption{Substitution}
        \label{fig:exa-online-cls-sub}
    \end{subfigure}
    \caption{Greedy compression configuration search on \pfntwo{}. We report the ratio of AUC$_{found}$ to the best possible AUC on the test set as well as overall performance degradation (blue). For regression results, see Figure~\ref{fig:exa-online-reg} in Appendix~\ref{app:x-plots}.}
    \label{fig:exa-online-cls}
\end{figure}

\begin{figure}[t]
    \centering
    \begin{subfigure}{0.32\linewidth}
        \includegraphics[width=\linewidth]{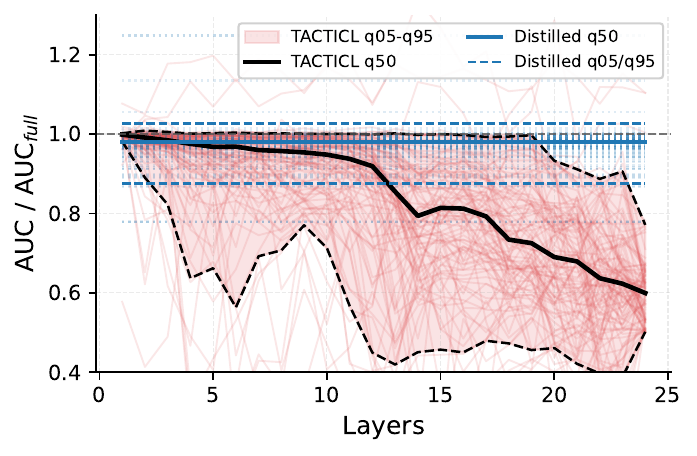}
        \caption{Sub-from-Last}
    \end{subfigure}
    \begin{subfigure}{0.32\linewidth}
        \includegraphics[width=\linewidth]{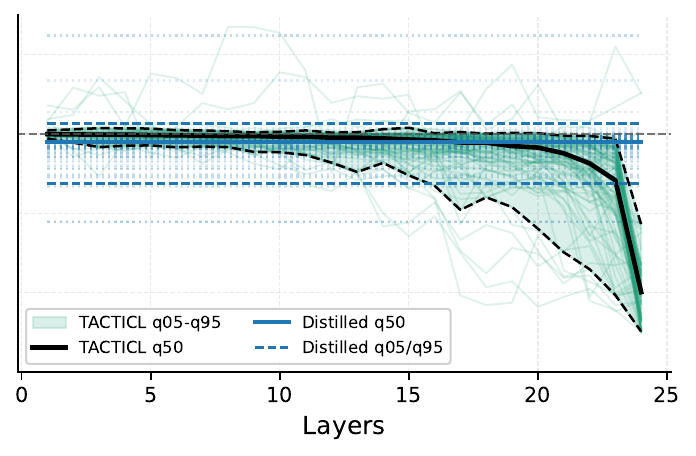}
        \caption{AUC-guided}
        \label{fig:scaling-pfn2.5-cls-sub-auc}
        
    \end{subfigure}
    \begin{subfigure}{0.32\linewidth}
        \includegraphics[width=\linewidth]{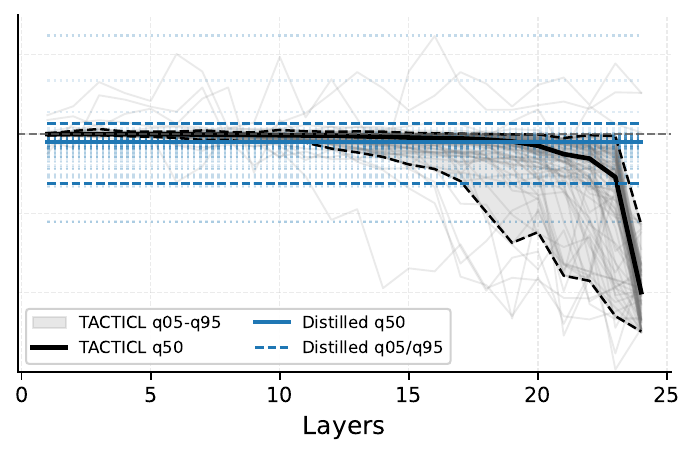}
        \caption{Stability-guided}
    \end{subfigure}
    \caption{\methodname{} substitution applied to \pfntwohalf{}. We report the ratio of performances in AUC aggregated across \tabarenac{} datasets for substituting layers across 3 strategies. Appendix~\ref{app:x-plots} provides results for deletion (\cref{fig:scaling-pfn2.5-cls-del}) as well as regression (Figures~\ref{fig:scaling-pfn2.5-reg-sub}, \ref{fig:scaling-pfn2.5-reg-del}).
    \label{fig:scaling-pfn2.5-cls-sub}}
\end{figure}

\textbf{Greedy Search Results.} To evaluate whether greedy search ends up in substantially worse local minima, we compare its performance to the optimal configuration identified in our proof-of-concept in Section~\ref{sec:poc}. In Figure~\ref{fig:exa-online-cls}, we compare the median relative drop in AUC with respect to the oracle performance, i.e., the optimal configuration with $n$ layers pruned, across all \tabarenac{} datasets. (see Figure~\ref{fig:exa-online-reg} in Appendix~\ref{app:x-plots} for regression results). Additionally, we report the ratio of the performance metrics between the full model and the oracle pruning performance. For layer deletion (left), performance degrades beyond six dropped layers, with a sharp drop and a substantial widening of variance across datasets. However, for layer substitution (right), greedy search recovers $>90$\% of the oracle AUC for configurations of up to 8 substituted layers and maintains this threshold for more than half of the datasets. This asymmetry further supports the use of adapter substitution over pure deletion: adapters not only restore local performance but also extend the range over which greedy search remains reliable. Similarly for regression, RMSE increases by less than $10\%$ for substituting up to 7 layers (and by less than $30\%$ for deleting up to 4 layers, respectively). Therefore, we conclude that the locality assumption holds for compressing substantial parts of the model and suffices for our algorithm.

\textbf{Scaling to \pfntwohalf{}.} 
Next, we evaluate a state-of-the-art model, \pfntwohalf{}, at scale. We focus on substituting layers for classification tasks using validation performance as a guiding metric. We report additional results on layer deletion and regression tasks in Appendix~\ref{app:x-plots}.
Additionally, we study two simple baselines. First, substitution or dropping layers starting from the last to first; and second, we distill the full model into a shallow MLP (see Appendix~\ref{app:distillation} for further details).

In Figure~\ref{fig:scaling-pfn2.5-cls-sub}, we compare full-model performance against the best configuration found by \methodname{} and against the distillation baseline. The results are consistent with our small-scale experiments: our combined pruning and adaptation allow us to replace up to $15$ layers with negligible performance degradation, and our method continues to outperform distillation up to $17$ layers, all while potentially preserving the model's in-context-learning capabilities. The drop/substitute-from-last baseline, by contrast, underperforms substantially compared to both \methodname{} and distillation, confirming that the strong performance of the last-layer heuristic on some architectures does not hold for \pfntwohalf{} and that a searched or learned selection of which layers to remove is necessary.

\structure{Show that the greedy search scales to TabPFN~v2.5, where exhaustive search would be infeasible; present results.}

\subsection{Can It Still Learn In-context? Retaining ICL Capabilities under Distribution Shift}
\label{sec:not-mlp}

\structure{Run \methodname\ on TabPFN~v2.5; evaluate the compressed model under distribution shift (uninformative features, permuted feature order, label flip) that would require re-running distillation to an MLP. Show that ICL capabilities are preserved.}

By substituting a layer with a shallow adapter network, we explicitly introduce in-weight learning (IWL) components and thus memorization. Since we keep several transformer layers identified as critical for context-reading and preserve their weights, the model's ICL ability should stay intact. Ideally, the model has been adapted to a downstream task but remains robust to mild distribution shifts that may occur in real-world tasks. To study this, we run two experiments. First, we evaluate the compressed model under three perturbations that alter the context without changing the underlying task. If the model can still learn in-context, then performance should stay the same. Second, we evaluate the model on datasets different from the one used for compression. If compression specializes a model for a task, we should observe a performance drop on other, unrelated tasks.

\begin{figure}[tb]
    \centering
    \begin{subfigure}{0.48\linewidth}
        \includegraphics[width=\linewidth]{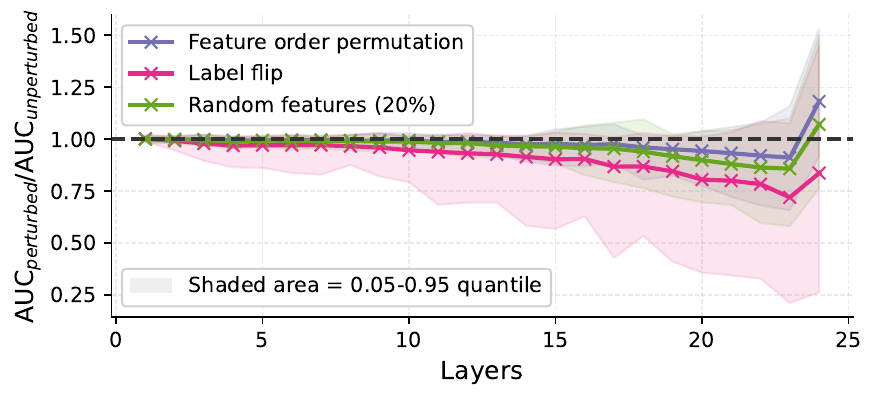}
        \caption{Perturbations evaluation}
        \label{fig:perturbation-cls}
    \end{subfigure}
    \begin{subfigure}{0.48\linewidth}
        \includegraphics[width=\linewidth]{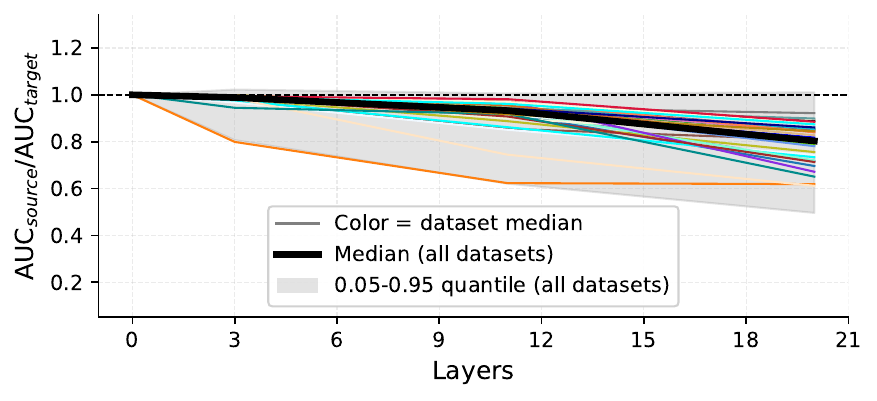}
        \caption{Cross-dataset Evaluation}
        \label{fig:cross-dataset-cls}
    \end{subfigure}
    \caption{Evaluations on perturbed datasets (OOD) (\ref{fig:perturbation-cls}) and cross-dataset generalization (evaluated on 3, 11, and 20 layers substituted) (\ref{fig:cross-dataset-cls}) confirm that the compressed model retains its ICL capabilities beyond the training distribution. We show results for regression in Figure~\ref{fig:ood-reg}}
    \label{fig:ood-cls}
\end{figure}
\textbf{Generalization under Perturbations.} We inject uninformative features, randomly permute the feature order, and flip labels. These perturbations would require a full re-distillation or training from scratch for an IWL approach, whereas any ICL model should handle them gracefully. We report the absolute difference to unperturbed performance in Figure~\ref{fig:perturbation-cls} (Figure~\ref{fig:perturbation-reg} for regression), showing that performance remains stable under perturbations. The smaller gap at higher compression can be attributed to overall worse performance, as seen already in Figure~\ref{fig:scaling-pfn2.5-cls-sub-auc} (Figure~\ref{fig:scaling-pfn2.5-reg-sub-rmse} for regression).

\textbf{Generalization across Datasets.} We apply the configuration found by \methodname{} on one dataset and evaluate it on a different dataset, comparing it to the performance of a compression configuration searched directly on that target dataset. Negative transfer in this cross-dataset setting confirms that \methodname{} specializes the model for the source task. We observe negative cross-dataset transfer in Figure~\ref{fig:cross-dataset-cls} (Figure~\ref{fig:cross-dataset-reg} for regression) for most compressed models, indicating that \methodname{} indeed adapts the model to a task by pruning away layers not needed for inference on the target task and by mitigating the resulting performance drop via the learned adapter.

\subsection{Do We Need Fine-tuning? Aligning Layers For Full Recovery}
\label{sec:finetuning}
\begin{figure}[b]
    \centering
    \includegraphics[width=\linewidth]{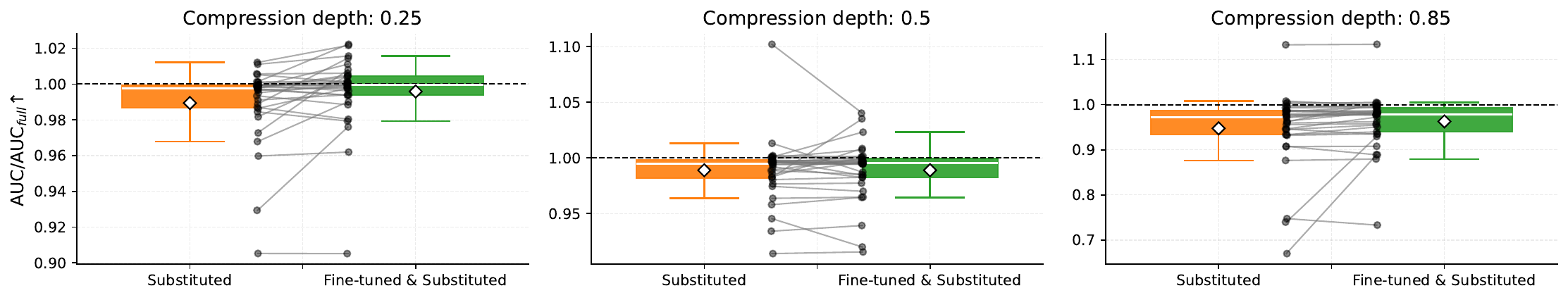}
    \caption{Change compared to the full model in AUC after fine-tuning the adapter weights at three compression levels. 
    }
    \label{fig:finetuning}
\end{figure}

\structure{Fine-tune the \adapter\ weights (keeping all other weights frozen) on the downstream task to encourage task-specific memorization while preserving ICL capabilities;            present results showing performance improvement.}

To recover performance lost during pruning, we evaluate the effect of fine-tuning the compressed model end-to-end. Crucially, we freeze all weights \emph{except} the adapter parameters. This is a deliberate design choice: we are interested in encoding task-specific information into the adapters rather than modifying the model's in-context inference circuitry. 
To prevent overfitting and maintain generalization, we fine-tune on a mixture of pretraining and in-domain data~\citep{bethune2025scaling} (see Appendix~\ref{app:finetunning} for more details and an ablation on the data mixture). We plot performance at different compression levels in Figure~\ref{fig:finetuning} and ~\ref{fig:finetuning-reg} and observe a general trend: fine-tuning improves performance and can potentially recover full model performance. This effect is more pronounced at lower compression rates, where we even observe improved performance for some datasets. 
\subsection{Does \methodname{} Speed Up Prediction?}
\label{sec:speedup}
\begin{figure}[t]
    \centering
    \begin{subfigure}{0.48\linewidth}
        \includegraphics[width=\linewidth]{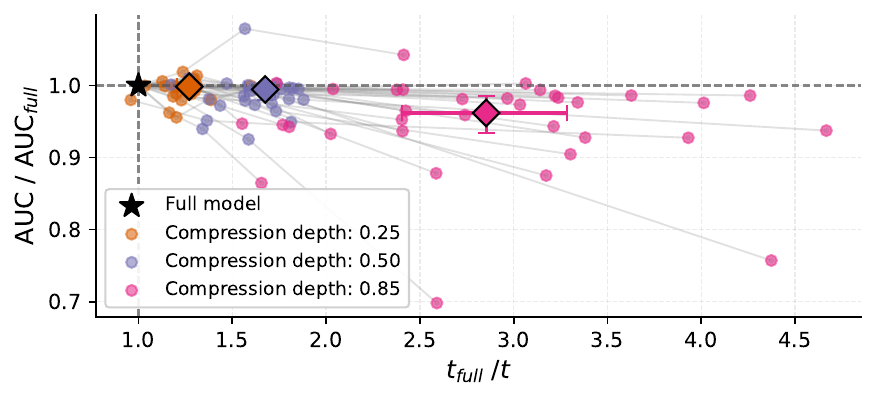}
    \caption{Classification}
        
    \end{subfigure}
    \begin{subfigure}{0.48\linewidth}
        \includegraphics[width=\linewidth]{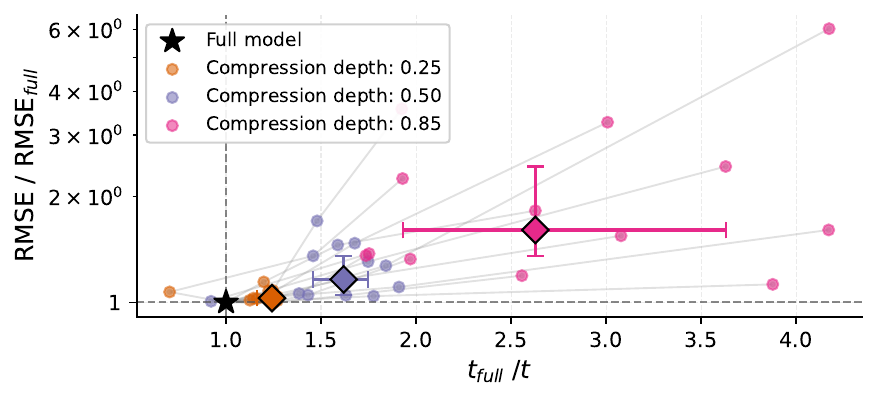}
        \caption{Regression}
    \end{subfigure}
    \caption{Ratios of \methodname{} speed-up vs task performance across three compression ratios on all evaluated datasets.}
    \label{fig:speedup}
\end{figure}

Finally, we verify that layer substitution speeds up inference, timing predictions directly on the held-out test set using the models selected by our greedy search at three compression budgets (25\%, 50\%, 85\% of transformer blocks substituted with a lightweight MLP). \Cref{fig:speedup} shows that speed-ups are substantial but sub-linear: for classification, inference is $1.25\times$, $1.63\times$, and $2.83\times$ faster at the three budgets, with essentially no AUC cost at 25--50\% (median ratio $\geq0.995$) and only a 4\% drop at 85\%. Regression shows similar speed-ups ($1.25\times$, $1.61\times$, $2.66\times$) but degrades more sharply, with RMSE increasing by roughly 3\%, 16\%, and 61\%---indicating that classification tolerates substituted representations far better than regression, so compression budgets should be applied more conservatively for regression tasks. We report inference costs for larger synthetic datasets in (\cref{fig:speedup-synth}), showing that speed-ups may reach $1.4\times$, $2.0\times$, and $5.4\times$ at the three budgets.

\section{Conclusion}\label{sec:conclusion}
We present \methodname{}, a framework for simultaneously pruning and adapting tabular foundation models to downstream tasks. \methodname{} efficiently searches over layer configurations to identify which layers to retain or drop. Applied to \pfntwohalf{}, our method achieves up to $5.4\times$ inference speedup, with an average of $2.78\times$, while maintaining competitive performance on our benchmark tasks. This opens a broader discussion on the trade-off between in-context and in-weights learning in tabular foundation models. Notably, we find that layer-importance metrics commonly used in the literature fail to identify good compression configurations. Furthermore, we found that no single pruning configuration is universally optimal, as the layers that matter most vary considerably from dataset to dataset. 

We further show that fine-tuning at low compression rates not only recovers performance lost to compression but can surpass that of the original, unpruned model. This suggests that \methodname{} not only reduces inference costs, but also serves as an efficient mechanism for domain adaptation. 
We consider several directions promising for future work: (1) hyperparameter optimization for the fine-tuning stage, (2) identifying additional suitable performance proxies beyond the cheap metrics that fall short here, and (3) natively building dataset-aware layer-dropping support into the model itself.

\textbf{Limitations.} The current framework does not yet provide a principled criterion for determining the tradeoff between compression level and performance. Instead, a user is provided with a substitution path and can decide which compressed model to finetune. Moreover, our empirical evaluation is limited to tasks within the pre-training constraints of \pfntwohalf{}{}. Here, we focus on establishing the feasibility and necessity of task-aware compression, and defer a larger-scale evaluation across additional tabular foundation models, larger input regimes, and the full \tabarenaname{} benchmark and leaderboard to future work.

\begin{acknowledgements}
This research has been funded by the Federal Ministry of Research, Technology and Space of Germany and the state of North Rhine-Westphalia as part of the Lamarr Institute for Machine Learning and Artificial Intelligence.
\end{acknowledgements}





\bibliography{bib/strings,bib/references,bib/lib,bib/proc,bib/myproc}
\newpage
\section*{Submission Checklist}


\begin{enumerate}
\item For all authors\dots
  \begin{enumerate}
  \item Do the main claims made in the abstract and introduction accurately
    reflect the paper's contributions and scope?
    \answerYes{}
  \item Did you describe the limitations of your work?
    \answerYes{} They are described in Section~\ref{sec:conclusion}.
  \item Did you discuss any potential negative societal impacts of your work?
    \answerNo{} Our work presents foundational research on compressing neural networks and we do not expect any negative societal impact that goes beyond a standard machine learning research paper and requires a dedicated discussion.
  \item Did you read the ethics review guidelines and ensure that your paper
    conforms to them? (see \url{https://2022.automl.cc/ethics-accessibility/})
    \answerYes{}
  \end{enumerate}
\item If you ran experiments\dots
  \begin{enumerate}
  \item Did you use the same evaluation protocol for all methods being compared (e.g.,
    same benchmarks, data (sub)sets, available resources, etc.)?
    \answerYes{}
  \item Did you specify all the necessary details of your evaluation (e.g., data splits,
    pre-processing, search spaces, hyperparameter tuning details and results, etc.)?
    \answerYes{}
  \item Did you repeat your experiments (e.g., across multiple random seeds or
    splits) to account for the impact of randomness in your methods or data?
    \answerYes{} We used 1 fold for Experiment \ref{sec:poc} and 3 folds for other experiments on \tabarena{} different datasets.
  \item Did you report the uncertainty of your results (e.g., the standard error
    across random seeds or splits)?
    \answerYes{} Over datasets and folds.
  \item Did you report the statistical significance of your results?
    \answerNo{} We perform exploratory research and not confirmatory research.
  \item Did you use enough repetitions, datasets, and/or benchmarks to support
    your claims?
    \answerYes{}
  \item Did you compare performance over time and describe how you selected the
    maximum runtime?
    \answerYes{} We ran our method until convergence (i.e., removing/replacing all layers) and analyzed intermediate and final results.
  \item Did you include the total amount of compute and the type of resources
    used (e.g., type of \textsc{gpu}s, internal cluster, or cloud provider)?
   \answerYes{} See Section~\ref{sec:experiments}.
  \item Did you run ablation studies to assess the impact of different
    components of your approach?
    \answerYes{}
  \end{enumerate}
\item With respect to the code used to obtain your results\dots
  \begin{enumerate}
\item Did you include the code, data, and instructions needed to reproduce the
    main experimental results, including all dependencies (e.g.,
    \texttt{requirements.txt} with explicit versions), random seeds, an instructive
    \texttt{README} with installation instructions, and execution commands
    (either in the supplemental material or as a \textsc{url})?
    \answerYes{} We provide code at \codeurl{}.
  \item Did you include a minimal example to replicate results on a small subset
    of the experiments or on toy data?
    \answerYes{}
  \item Did you ensure sufficient code quality and documentation so that someone else
    can execute and understand your code?
    \answerYes{}
  \item Did you include the raw results of running your experiments with the given
    code, data, and instructions?
    \answerNo{}
  \item Did you include the code, additional data, and instructions needed to generate
    the figures and tables in your paper based on the raw results?
    \answerNo{} We did not upload the pretraining data for finetuning (Section~\ref{sec:finetuning} and expect that running the script will lead to slightly different results. We will provide the data when the code is released.
  \end{enumerate}
\item If you used existing assets (e.g., code, data, models)\dots
  \begin{enumerate}
  \item Did you cite the creators of used assets?
    \answerYes{}
  \item Did you discuss whether and how consent was obtained from people whose
    data you're using/curating if the license requires it?
   \answerNA{}
  \item Did you discuss whether the data you are using/curating contains
    personally identifiable information or offensive content?
    \answerNA{}
  \end{enumerate}
\item If you created/released new assets (e.g., code, data, models)\dots
  \begin{enumerate}
    \item Did you mention the license of the new assets (e.g., as part of your
    code submission)?
    \answerNo{} We will release the code under an OSI-approved license upon acceptance.
    \item Did you include the new assets either in the supplemental material or as
    a \textsc{url} (to, e.g., GitHub or Hugging Face)?
    \answerYes{} We provide new assets at \codeurl{}.
  \end{enumerate}
\item If you used crowdsourcing or conducted research with human subjects\dots
  \begin{enumerate}
  \item Did you include the full text of instructions given to participants and
    screenshots, if applicable?
    \answerNA{}
  \item Did you describe any potential participant risks, with links to
    institutional review board (\textsc{irb}) approvals, if applicable?
    \answerNA{}
  \item Did you include the estimated hourly wage paid to participants and the
    total amount spent on participant compensation?
    \answerNA{}
  \end{enumerate}
\item If you included theoretical results\dots
  \begin{enumerate}
  \item Did you state the full set of assumptions of all theoretical results?
    \answerNA{}
  \item Did you include complete proofs of all theoretical results?
    \answerNA{}
  \end{enumerate}
\end{enumerate}

\newpage

\appendix

\section{Metrics}
\paragraph{AUC} For classification tasks we report the area under the receiver operating characteristic curve (AUC). For a binary task with predicted scores $\hat{p}_i$ and true labels $y_i \in \{0, 1\}$, the ROC curve traces the true positive rate
\[
\mathrm{TPR}(\tau) = \frac{|\{i : y_i = 1,\ \hat{p}_i \geq \tau\}|}{|\{i : y_i = 1\}|}
\]
against the false positive rate
\[
\mathrm{FPR}(\tau) = \frac{|\{i : y_i = 0,\ \hat{p}_i \geq \tau\}|}{|\{i : y_i = 0\}|}
\]
as the decision threshold $\tau$ is swept over $[0, 1]$, and AUC is the area under this curve:
\[
\mathrm{AUC} = \int_0^1 \mathrm{TPR}\big(\mathrm{FPR}^{-1}(u)\big) \, du .
\]
Equivalently, AUC is the probability that a randomly chosen positive example is ranked above a randomly chosen negative example:
\[
\mathrm{AUC} = \Pr\big(\hat{p}_{i^+} > \hat{p}_{i^-}\big), \qquad y_{i^+} = 1,\ y_{i^-} = 0,
\]
with ties counted as one-half. For multiclass tasks, we compute the one-vs-one (OvO) pairwise average of this quantity over all class pairs.

\paragraph{RMSE} For regression tasks we report the root mean squared error between predictions $\hat{y}_i$ and targets $y_i$ over $n$ evaluation examples:
\[
\mathrm{RMSE} = \sqrt{\frac{1}{n}\sum_{i=1}^{n}\big(\hat{y}_i - y_i\big)^2} .
\]

\section{Shapley Values of Layers}
\label{app:shapley}
\begin{figure}[b]
    \centering
    \begin{subfigure}{0.49\linewidth}
        \includegraphics[width=\linewidth]{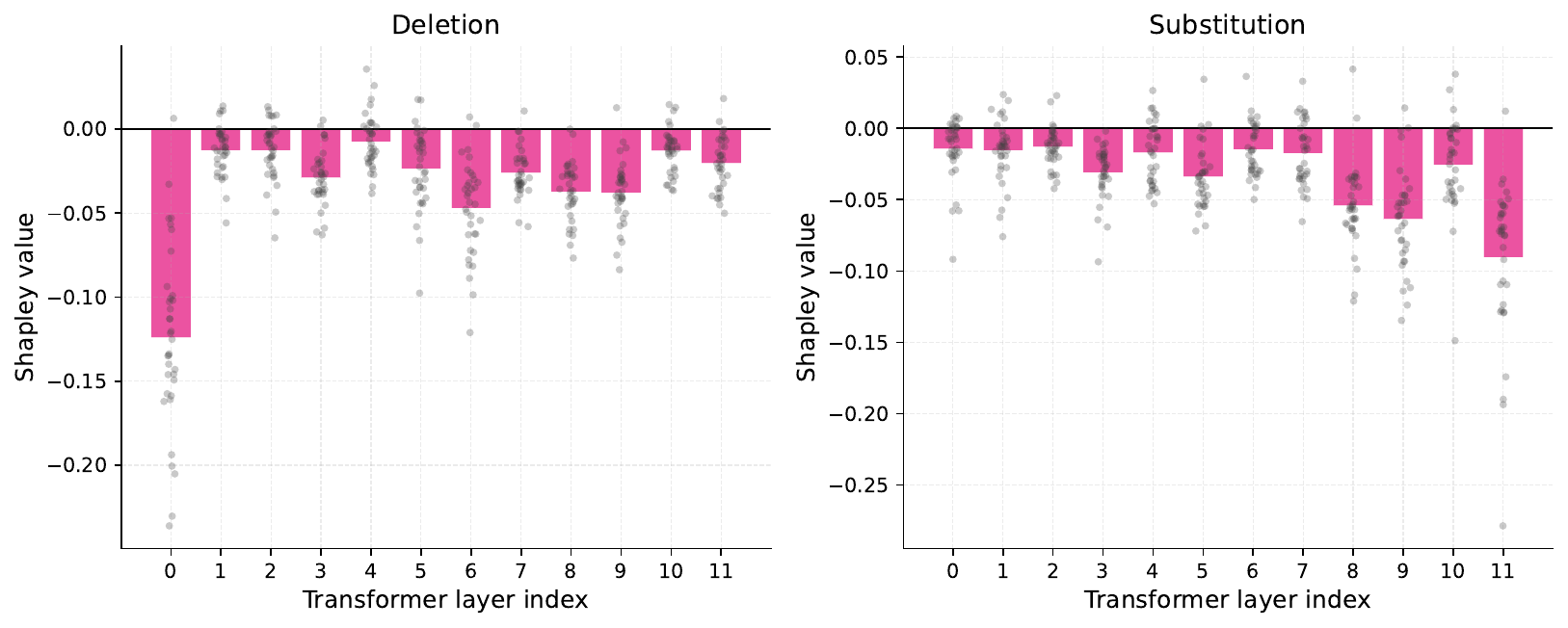}
        \caption{Classification}
        \label{fig:shapley-cls}
    \end{subfigure}
    \begin{subfigure}{0.49\linewidth}
        \includegraphics[width=\linewidth]{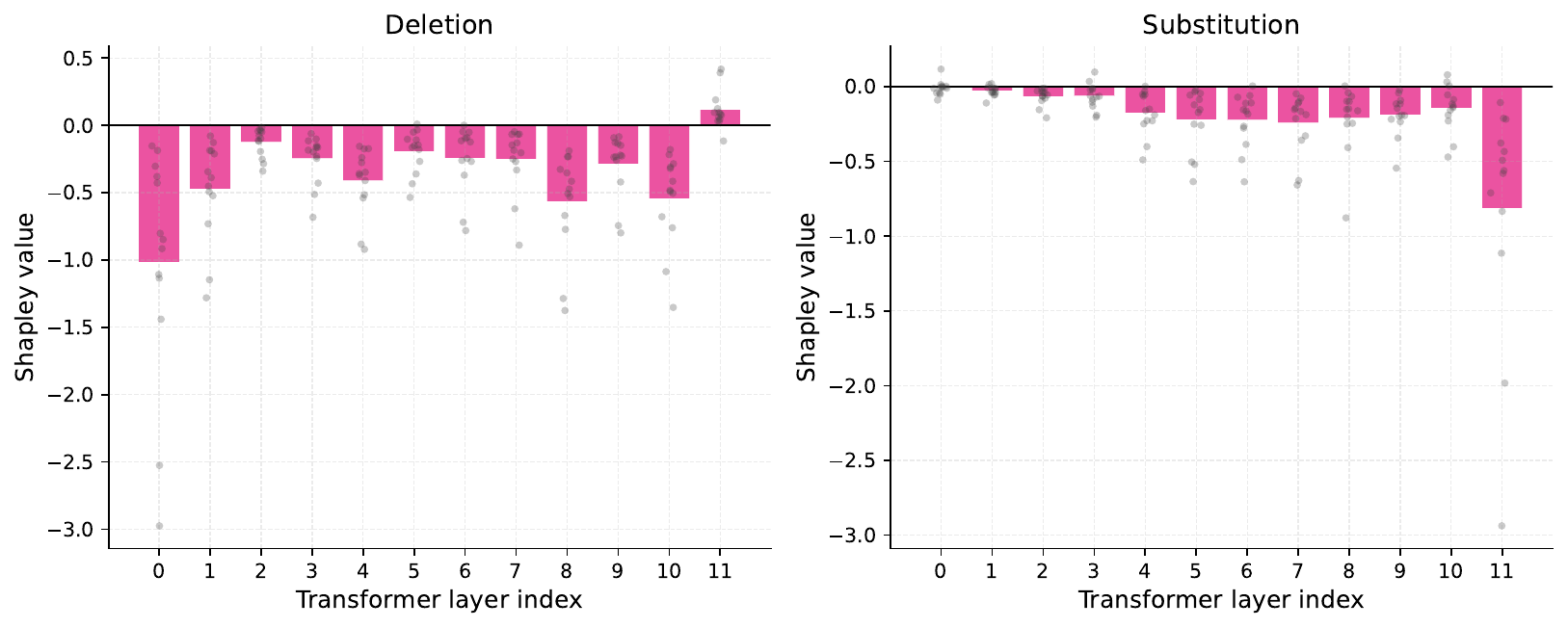}
        \caption{Regression}
            \label{fig:shapley-reg}
    \end{subfigure}
    \caption{Shapley across layer deletions/substitution based on the exhaustive evaluation of the \pfntwo{} over the \tabarena{} \tabarenaname{} datasets. The lower the value, the more harmful the change applied to the layer. Individual point -- dataset; bar -- mean over all datasets. As can be seen from the individual points, the variance over datasets is high, meaning no universal order exists for deleting/substituting the layers.}
    \label{fig:shapley}
\end{figure}
For each task type, we treat the $N$ transformer layers as players in a cooperative game, where a coalition is any subset of layers dropped from the compressed/pruned model, and the characteristic function $v(S)$ is the mean value of the task's native metric (AUC for classification, negative RMSE for regression) for the configurations whose layer-drop mask matches $S$.  Since the underlying sweep exhaustively evaluates all $2^N$ possible layer-inclusion masks with one configuration per mask, this value function can be read off directly by indexing runs by their binary layer mask. The exact Shapley value for layer $i$ is then obtained via the standard closed-form expression, $\phi_i = \sum_{S \subseteq N\setminus{i}} \frac{|S|!(N-|S|-1)!}{N!}\big(v(S\cup{i}) - v(S)\big)$, summing the layer's marginal contribution over every possible coalition of the other layers, weighted by the combinatorial term that averages uniformly over all orderings in which the layer could be added. This is computed per dataset-fold and then averaged across datasets to obtain the per-layer Shapley attribution reported for each metric, giving a fair, order-independent measure of each layer's contribution to overall model quality. 

Shapley values of \pfntwo{} obtained on the results of exhaustive evaluation from Section~\ref{sec:poc} are presented in \cref{fig:shapley}. These values could serve as a compression roadmap if they aligned across datasets: less-negative values identify layers that are safe or even beneficial to remove. Under substitution, layers 0-2 (0-3 for the regression) seem to be safe to compress, while layer~11 should be avoided. Under deletion, layer~0 carries a strongly negative Shapley value and is the dominant constraint---excluding it is the single most important rule for staying within the positive-performance region, which is consistent with the findings of \citet{balef2025towards}.

\section{Datasets}
\label{app:datasets}
\subsection{Evaluation Datasets}

For the exhaustive and online search evaluations, we use datasets drawn from the TabArena benchmark suite (OpenML study/suite 457), a curated collection of tabular classification and regression tasks. From TabArena's datasets, we retain only those that fit within TabPFN v2.5's native pretraining limits (at most 50{,}000 training samples and 2{,}000 features, checked against the actual per-fold training-split size) without requiring the \texttt{ignore\_pretraining\_limits} override, yielding a fixed suite of \textbf{\tabarena{} datasets} --- \tabarenac{} classification and \tabarenar{} regression --- used consistently across both search procedures.

The evaluations with \pfntwo{} done on all \tabarena{} datasets on a single fold (fold 0) due to time constraints (larger dataset taking more than a day to evaluate). The evaluations with \pfntwohalf{}  use the same \tabarena{} datasets across 3 folds (folds 0--2) to assess robustness of the discovered layer-drop paths to the train/test split. Both use the full available train and test splits (no subsampling), a fixed data seed and model seed, and no random feature perturbation, for direct comparability between the two search procedures and with the downstream cross-dataset, perturbation, and inference-timing evaluations that build on their outputs.
\subsection{Perturbations}

To test whether the compressed models discovered by the greedy online search remain robust under distribution shift, we re-evaluate the previously selected layer-drop paths (one per dataset/fold 
) 
on perturbed versions of the test data. 
Three perturbations are applied, deterministically seeded for reproducibility:

\begin{itemize}
    \item \textbf{Feature order permutation}: the columns of $X$ are randomly permuted (a single random permutation drawn per experiment and applied consistently to both the train and test splits), testing sensitivity to feature ordering.
    \item \textbf{Random features (+20\%)}: uninformative features drawn i.i.d.\ from $\mathrm{Uniform}(0,1)$ are appended to $X$, in a number equal to 20\% of the original feature count, testing robustness to the presence of irrelevant/noisy features.
    \item \textbf{Label flip} (classification only): class labels are remapped through a fixed-point-free permutation (a derangement) of the label set --- for binary tasks this simply swaps the two classes --- applied identically to train and test labels, testing whether the model's behavior is tied to the specific label semantics rather than to structure in $X$.
\end{itemize}

Each selected path is evaluated once under each applicable perturbation (all three for classification tasks, the two feature-based perturbations for regression tasks), and the resulting test metric is compared against the same path's performance on the original, unperturbed data to quantify robustness of the discovered compressed configurations.

\subsection{Synthetic Prior Data Injection}
\label{sec:prior_injection}
To prevent the fine-tuned adapters from drifting away from the general-purpose behavior learned during TabPFN's original pretraining, we periodically interleave ordinary downstream fine-tuning epochs with epochs trained on synthetic data sampled from TabPFN's own pretraining prior, rather than the target dataset. Every $N$ epochs (a configurable cadence), the training batch is replaced with one or more precomputed synthetic "prior" batches instead of real data.

Synthetic data is generated offline using TabICL's \citep{tabicl2} open-source structural-causal-model-based prior generator (\texttt{graph\_scm}), the same class of generative process used to pretrain TabPFN models. Each precomputed batch contains 256 synthetic datasets with randomly varying properties: number of features (2--100), sequence length (up to 1024 rows), number of classes for classification (up to 10), and train/test split fraction (drawn between 10\% and 90\% of the sequence length). 
At injection time, each synthetic sample is passed through the compressed model's preprocessing pipeline used for real data, preserving its own generator-assigned train/test split boundary, so that it is represented with the expected preprocessing.

\section{Additional Details on Adapter Training}
\label{app:adapter-training}
To compensate for a dropped transformer layer, we train a small MLP adapter to reconstruct the frozen teacher's hidden state after that layer from its hidden state before the layer, using the same input/output dimensionality as the model's hidden size — one adapter is trained per dropped layer. Each adapter is a lightweight two-layer MLP (hidden width 128) with GELU activations and a skip connection, trained independently and from scratch. Training minimizes mean squared error between the adapter's output and the dropped layer's true output, using AdamW (lr = 4e-3, weight decay $1.5\times10^{-5}$) with a linear warmup followed by a cosine-annealing-with-warm-restarts schedule, for up to 2000 epochs with early stopping (patience of 150 epochs) on a held-out validation split of the cached hidden states. All adapters are pretrained and cached offline prior to the configuration search, and the training data is described in Figure~\ref{fig:overview}, decoupling adapter training from the pruning search and subsequent end-to-end fine-tuning stages.

\section{Adapter Architecture Search}
\label{app:adapter-nas}

We evaluated a set of $10$ adapter architectures varying in model size and activation function on a held-out set of five TALENT~\citep{talent} classification datasets (\textit{golf\_play\_dataset\_extended}, \textit{Basketball\_c}, \textit{Customer\_Personality\_Analysis}, \textit{steel\_plates\_faults}, and \textit{dry\_bean\_dataset}) disjoint from both the main OpenML evaluation benchmark and the datasets used elsewhere in the pipeline. We report average AUC in Table~\ref{tab:adapter_grid} and use the best-performing architecture $C3\_wide(128)$ throughout the experiments in the main paper.

\begin{table}[h]
\centering
\caption{Adapter design grid used in the architecture ablation, ranked by mean downstream AUC across the 5-dataset evaluation panel.}
\label{tab:adapter_grid}
\begin{tabular}{lccccc}
\toprule
Config & Hidden width & Depth & Activation & Skip & Mean AUC \\
\midrule
\textbf{C3\_wide128} (deployed) & 128 & 2 & GELU     & Yes & \textbf{0.865} \\
E0\_MLP                         & 256 & 4 & ReLU     & No  & 0.822 \\
D1\_deep3                       & 64  & 3 & GELU     & Yes & 0.813 \\
C2\_wide64                      & 64  & 2 & GELU     & Yes & 0.810 \\
C1\_pure\_linear                & 0   & 1 & Identity & Yes & 0.786 \\
B3\_relu                        & 32  & 2 & ReLU     & Yes & 0.738 \\
A1\_no\_skip                    & 32  & 2 & SiLU     & No  & 0.714 \\
D2\_narrow\_deep                & 32  & 3 & GELU     & Yes & 0.693 \\
B2\_identity                    & 32  & 2 & Identity & Yes & 0.663 \\
B1\_gelu                        & 32  & 2 & GELU     & Yes & 0.629 \\
baseline                        & 32  & 2 & SiLU     & Yes & 0.575 \\
\bottomrule
\end{tabular}
\end{table}

\section{Distillation baseline}
\label{app:distillation}
As an additional point of comparison to our depth-pruning approach, we include a knowledge-distillation baseline that compresses TabPFN v2.5 by training a compact MLP student to reproduce the predictions of the full teacher model, rather than by removing internal layers. The distillation implementation is adapted from the open-source TabTune\footnote{\url{https://github.com/Lexsi-Labs/TabTune}} library \citep{tanna2026exploring}, simplified and consolidated for our teacher-student evaluation pipeline.

For each dataset and fold, we first fit the full TabPFN v2.5 model (the teacher) on the training split and record its predictions on that data as soft targets. We follow the original protocol of \citet{tanna2026exploring}. To account for varying teacher confidence across examples, we scale the distillation temperature per sample based on the entropy of the teacher's prediction, softening targets more for uncertain examples, and reweight the training loss with a bell-shaped confidence weighting that emphasizes examples of intermediate confidence. Table~\ref{tab:distill_hparams} lists the hyperparameters used.

\begin{table}[h]
\centering
\caption{Hyperparameters used for the distillation baseline.}
\label{tab:distill_hparams}
\begin{tabular}{ll}
\toprule
Hyperparameter & Value \\
\midrule
Base distillation temperature $T$        & 3.0 \\
Adaptive temperature range               & $[0.5T, 2T]$ \\
Adaptive temperature                     & enabled \\
Confidence weighting                     & enabled \\
KD/hard-label blend $\alpha$              & 0.7 (KD term), $1-\alpha$ (hard-label term) \\
Student architecture                     & residual MLP (hidden width 128, depth 2, GELU) \\
Optimizer                                & AdamW \\
Learning rate                            & $5 \times 10^{-4}$ \\
Weight decay                             & $1 \times 10^{-4}$ \\
Batch size                               & 256 \\
Dropout                                  & 0.1 \\
Max epochs                               & 2000 \\
Early stopping patience                  & 30 epochs \\
Teacher ensemble size ($n_\text{estimators}$) & 1 \\
\bottomrule
\end{tabular}
\end{table}

The student is trained with a combined objective that blends a distillation loss against the teacher's soft targets with a standard supervised loss against the ground-truth labels ($\alpha=0.7$ weight on the distillation term, $1-\alpha$ on the hard-label term), allowing the student to benefit from the teacher's learned decision boundary while remaining anchored to the true labels.

After training, we evaluate both the teacher and the distilled student on held-out test splits, reporting AUC for classification tasks and RMSE for regression tasks, along with the resulting performance gap. This gap serves as the reference point against which we measure how well our layer-pruned models preserve teacher performance at comparable compression levels.

\section{Proxy metrics}
\label{app:proxies}
Since the number of configurations is $\binom{L}{n}$, exhaustive search is generally intractable for large $L$, motivating the need for efficient proxies that estimate the performance of the healed model without explicit retraining. We consider several such proxies.

The first is \emph{embedding similarity} \citep{sajjad2023effect}, in our case cosine similarity, between the input and output hidden states of a layer. High similarity suggests that the layer induces only a minor transformation and is therefore easier to approximate with a lightweight substitute.

The second is \emph{direct performance} of other \methodname{} variants: the task performance of the model obtained by either removing or substituting a candidate set of layers $c$, evaluated directly (AUC for classification, RMSE for regression). We use each variant as a proxy for the other: if direct deletion performance predicts substitution performance, or vice versa, a single search would suffice for both deletion and substitution decisions, rather than requiring a separate search for each.

A third candidate metric is the \emph{approximation error} of a fixed-capacity model, such as an MLP, trained to replicate the layer's input--output mapping, using MSE loss. Layers that are harder to approximate accurately are likely more critical to preserve, making this error a natural indicator of which layers are most beneficial to retain.

Additionally, following \citet{balef2025towards} we calculate the \textit{sum of residual contributions} as a proxy. For a candidate set of layers to drop $\mathcal{D} = \{i_1,\dots,i_n\}$, we measure how much each dropped layer perturbs the residual stream relative to the running hidden-state magnitude,
\begin{equation}
\rho_i = \frac{\lVert h_i - h_{i-1} \rVert}{\lVert h_i \rVert},
\end{equation}
computed once from the original, unpruned model's cached hidden states $h_0, \dots, h_L$, and aggregate as $\mathcal{R}(\mathcal{D}) = \sum_{i \in \mathcal{D}} \rho_i.$
Layers that leave the residual stream nearly unchanged ($\rho_i \approx 0$) contribute little new information at that depth, in line with prior observations that deeper transformer layers apply near-redundant transformations to the residual stream~\citep{gromov2025unreasonable, men2025shortgpt}. A low $\mathcal{R}(\mathcal{D})$ therefore serves as a cheap proxy for "safe to remove".

Finally, we explore the \emph{stability} metric, introduced by \citet{chen2025streamlining}. 
Unlike other metrics, it is not cheap because it requires inference with the pruned model. 
Originally, it was introduced for generative multiple-choice QA, where the model yields only a perplexity score per answer choice rather than a full probability distribution. 
There, a sample is deemed stable if the model's correctness relative to ground truth (right/wrong) is unchanged after pruning, weighted by the original model's confidence. This discards substantial information: correctness-consistency alone cannot register how the underlying distribution moved even when the correct choice is preserved, meaning a pruned model that keeps the correct top choice while collapsing toward a near-uniform distribution over the remaining options registers as perfectly stable, despite being brittle to minor input perturbations, since only the \emph{original} model's confidence enters the weighting.

We adapt the metric to our setting, where we have access to the model's full predictive distribution rather than a per-choice perplexity score, allowing us to measure how much the entire distribution shifts, not just whether correctness is preserved. Concretely, for each sample $x$ we compute
\begin{equation}
s(x) = \exp\!\left(-D_{\mathrm{KL}}(p_{\text{orig}} \,\|\, p_{\text{pruned}})\right) \cdot m_{\text{orig}},
\end{equation}
where $m_{\text{orig}} = p^{(1)}_{\text{orig}} - p^{(2)}_{\text{orig}}$ is the original model's top-2 confidence margin, and aggregate as
\begin{equation}
\mathcal{S}_{\mathrm{KL}} = \frac{\sum_{x} s(x)}{\sum_{x} m_{\text{orig}}(x)}.
\end{equation}
For classification, $p_{\text{orig}}$ and $p_{\text{pruned}}$ are the models' predicted class-probability vectors. For the regression PFN, we exploit the fact that \tabpfn{} represents a continuous target not as a point estimate but as a categorical distribution over a fixed set of discretized bins covering the target range (with two unbounded tail bins for out-of-range support). We extract this per-bucket distribution from the model's logits and apply Equation~(1)--(2) unchanged, treating the bins as classes: $p_{\text{orig}}$ and $p_{\text{pruned}}$ are the softmax-normalized bin probabilities, and $m_{\text{orig}}$ is the margin between the top two bins. This lets us apply the same distributional stability measure to regression without any change to the underlying formula --- only the semantics of what constitutes a "class" changes, from label to target-range bucket.

\section{Additional Details on Fine-Tuning}
\label{app:finetunning}

After adapters are pretrained offline to reconstruct the dropped layers' outputs (Section~\ref{sec:adapter_training}), we further refine them end-to-end on the actual downstream task, for each of the depth fractions $\{0.25, 0.5, 0.85\}$ along the selected compression path. All backbone TabPFN parameters remain frozen; only the layer-substitution adapter MLPs are updated. Fine-tuning optimizes the model's real downstream objective rather than the reconstruction proxy: cross-entropy on TabPFN's output logits for classification, or the negative log-likelihood under TabPFN's own bar-distribution (histogram) output head for regression.

We use TabPFN's native meta-learning-style batched training interface: at each epoch, the training data is freshly re-split into an in-context "training" and "query" portion (a new random split each epoch, with a fixed 10\% validation holdout reserved separately), preprocessed through the model's standard pipeline, and used for a single differentiable forward/backward pass through the frozen backbone into the adapters, with the model restricted to a single estimator ($n_\text{estimators}=1$) during training (a requirement for gradient-based adapter injection). Table~\ref{tab:finetune_hparams} lists the hyperparameters used.

\begin{table}[h]
\centering
\caption{Hyperparameters used for adapter fine-tuning.}
\label{tab:finetune_hparams}
\begin{tabular}{ll}
\toprule
Hyperparameter & Value \\
\midrule
Optimizer                          & AdamW \\
Learning rate                      & $1 \times 10^{-5}$ \\
Weight decay                       & $1 \times 10^{-2}$ \\
LR schedule                        & linear warmup $\to$ cosine annealing \\
Warmup epochs                      & $\min(50, \text{epochs}/20)$ \\
Max epochs                         & 300 \\
Early stopping patience            & 150 epochs \\
Validation split ratio             & 0.1 \\
Checkpoint/early-stop metric       & AUC (classification) / RMSE (regression) \\
Prior injection cadence            & every 10 epochs \\
Injected batches per prior epoch   & 2 \\
\bottomrule
\end{tabular}
\end{table}

\begin{figure}[h]
    \centering
    \includegraphics[width=0.9\linewidth]{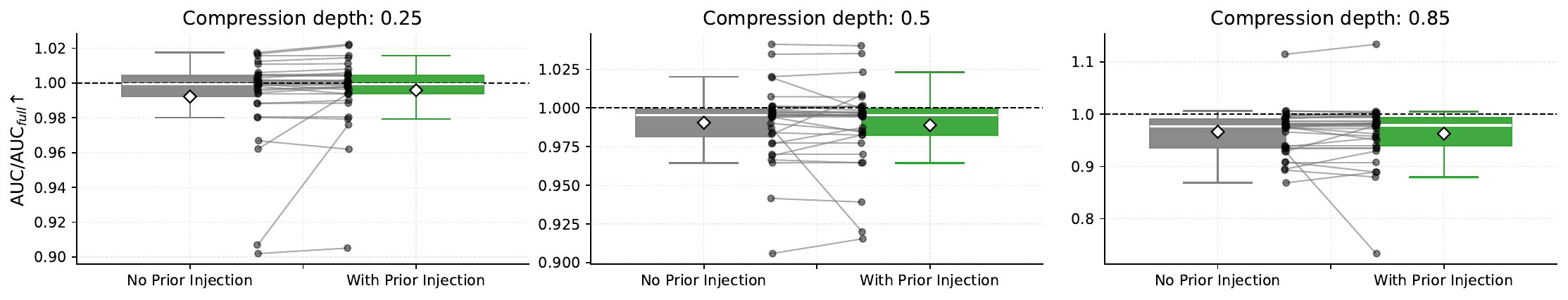}
    \caption{Change compared to the full model in AUC after fine-tuning the adapter weights at three compression levels              (25\%, 50\%, 85\% of layers replaced).}
    \label{fig:finetuning-cls-prior}
\end{figure}

\begin{figure}[h]
    \centering
    \includegraphics[width=0.9\linewidth]{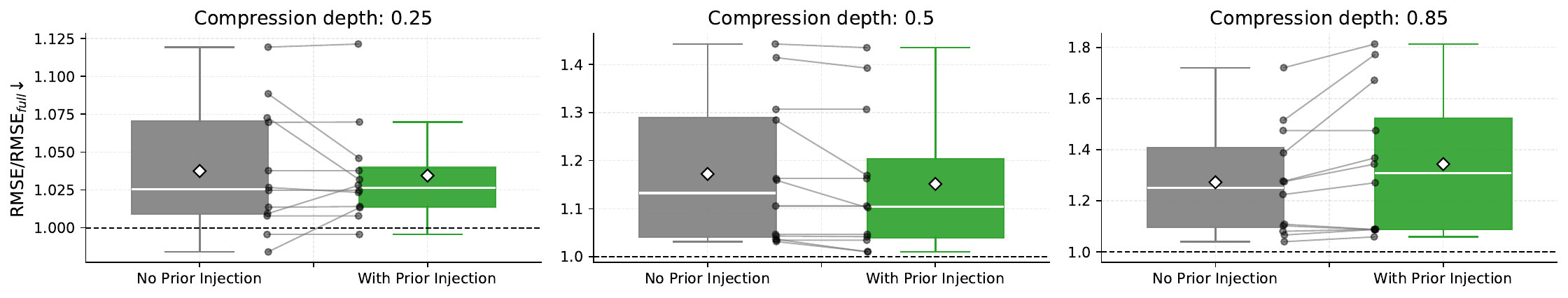}
    \caption{Change compared to the full model in AUC after fine-tuning the adapter weights at three compression levels              (25\%, 50\%, 85\% of layers replaced).}
    \label{fig:finetuning-reg-prior}
\end{figure}

Optimization uses AdamW with a linear warmup followed by a cosine-annealing learning-rate schedule, and early stopping on the held-out validation split (patience-based, tracking AUC for classification or RMSE for regression), restoring the best adapter checkpoint at the end of training. Every 10 epochs, the training batch is replaced with 2 precomputed synthetic prior batches instead of real data (Section~\ref{sec:prior_injection}), interleaving downstream fine-tuning with periodic re-exposure to TabPFN's pretraining-like distribution; we also run a matched no-injection control (identical hyperparameters, prior injection disabled) as an ablation (Fig.~\ref{fig:finetuning-cls-prior}, Fig.~\ref{fig:finetuning-reg-prior}), but results are not conclusive. We have realized that an experiment that assesses the ICL performance of the fine-tuned model might shine more light on the benefits of the prior injection, and leave this to the future work.
\begin{figure}[h]
    \centering
    \includegraphics[width=0.9\linewidth]{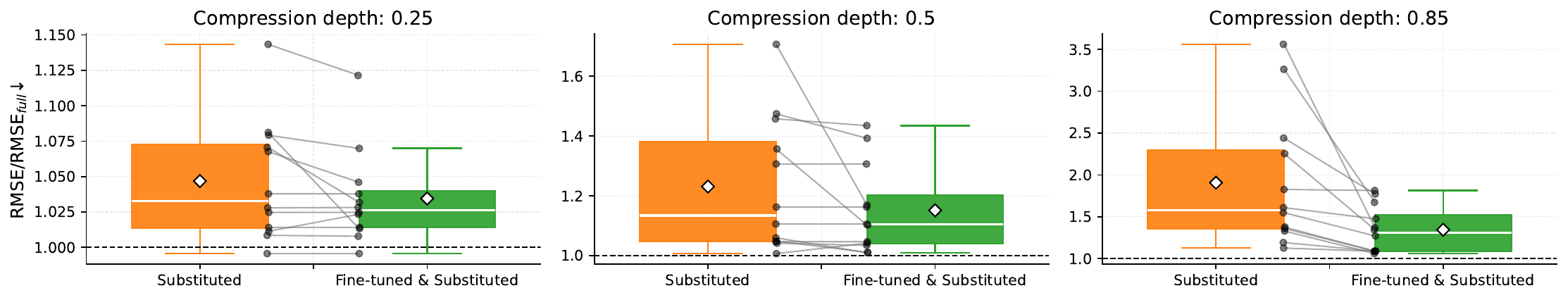}
    \caption{Change compared to the full model in AUC after fine-tuning the adapter weights at three compression levels              (25\%, 50\%, 85\% of layers replaced).}
    \label{fig:finetuning-reg}
\end{figure}

\newpage
\section{Additional plots}

\label{app:x-plots}
\begin{figure}[h]
    \centering
        \begin{subfigure}{0.4\linewidth}
        \includegraphics[width=\linewidth]{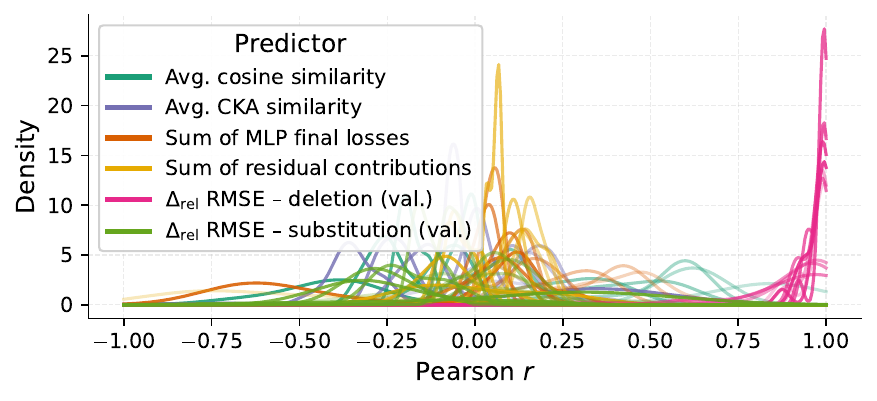}
        \caption{Deletion}
    \end{subfigure}
    \begin{subfigure}{0.4\linewidth}
        \includegraphics[width=\linewidth]{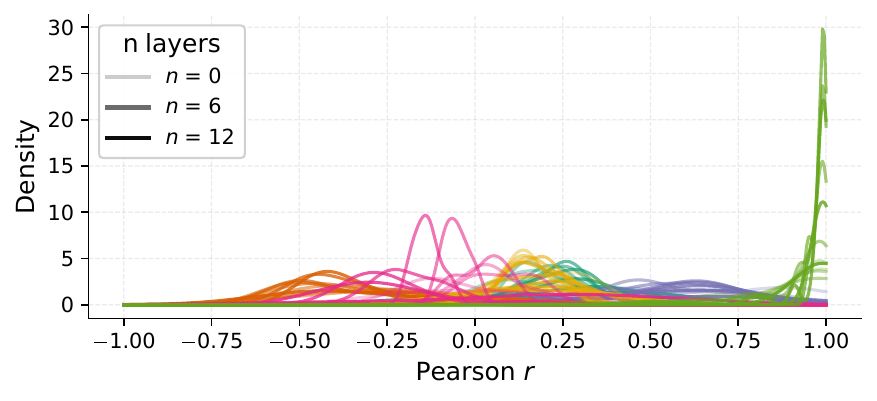}
        \caption{Substitution}
    \end{subfigure}
      \caption{Correlation between proxy scores and performance change for deleting (left) and substituting (right) $n$ layers for regression. None of the proxies reliably predict the performance impact.}
    \label{fig:proxies-vs-RMSE}
\end{figure}
\begin{figure}[h]
    \centering
        \centering
        \begin{subfigure}{0.4\linewidth}
        \includegraphics[width=\linewidth]{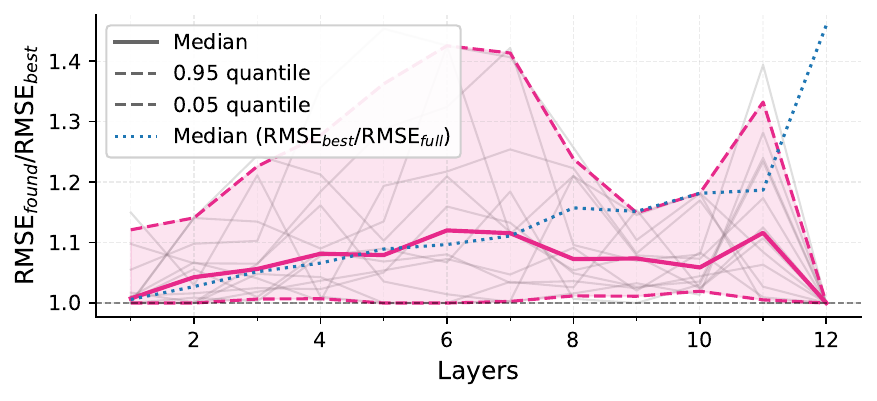}
        \caption{Deletion}
        \label{fig:exa-online-reg-del}
    \end{subfigure}
    \begin{subfigure}{0.4\linewidth}
        \includegraphics[width=\linewidth]{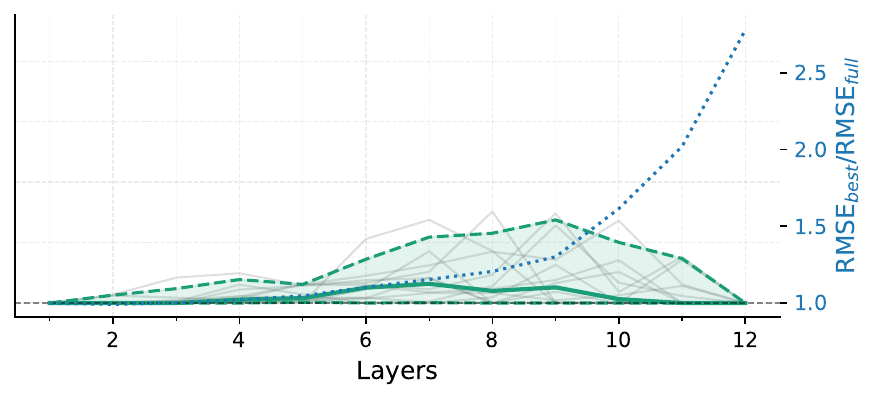}
        \caption{Substitution}
        \label{exa-online-reg-sub}
    \end{subfigure}
    \caption{Greedy search on TabPFN~v2: ratio of RMSE found by the greedy configuration over the oracle RMSE on the test set and overall performance (blue).}
    \label{fig:exa-online-reg}
\end{figure}
\begin{figure}[t]
    \centering
    \begin{subfigure}{0.45\linewidth}
        \includegraphics[width=\linewidth]{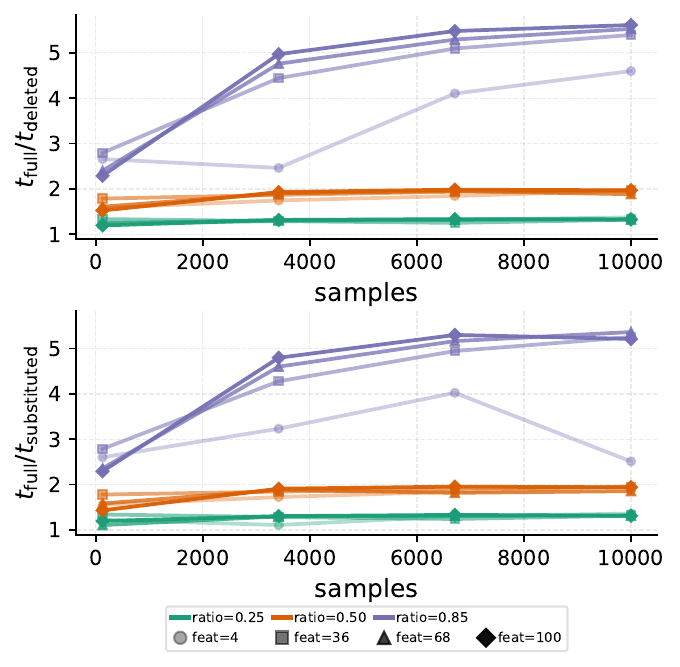}
        \caption{Per sample size}
    \end{subfigure}
    \begin{subfigure}{0.45\linewidth}
        \includegraphics[width=\linewidth]{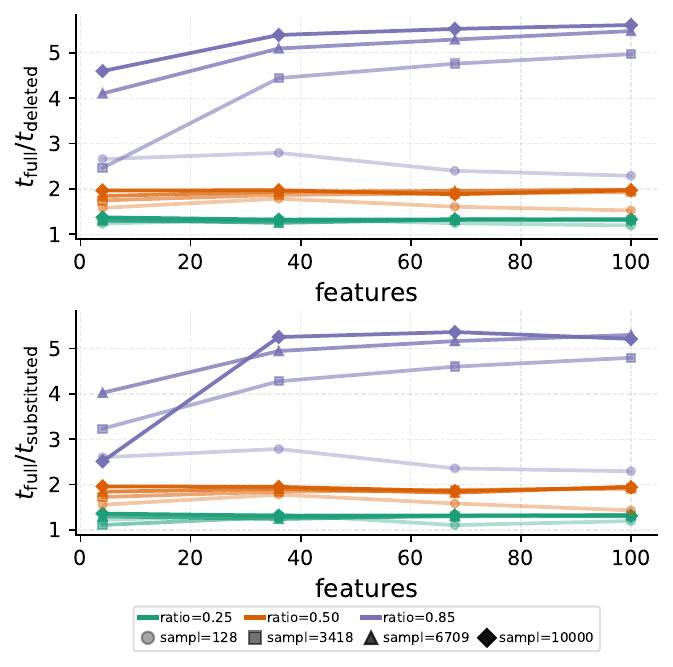}
        \caption{Per feature size}
    \end{subfigure}
    \caption{Speedup of \methodname{} for $0.25$, $0.5$ and $0.85$ compression ratios measured on the random synthetic datasets (classification task) of varying sample and feature sizes. The top row displays compression with deletion, bottom -- substitution. As evident from the plots, the overhead for substitution is minimal compared to deletion. The larger the dataset, the bigger the speedup -- up to $5.6\times$ for deletion and $5.4\times$ for substitution.  }
    \label{fig:speedup-synth}
\end{figure}
\begin{figure}[h]
    \centering
    \begin{subfigure}{0.3\linewidth}
        \includegraphics[width=\linewidth]{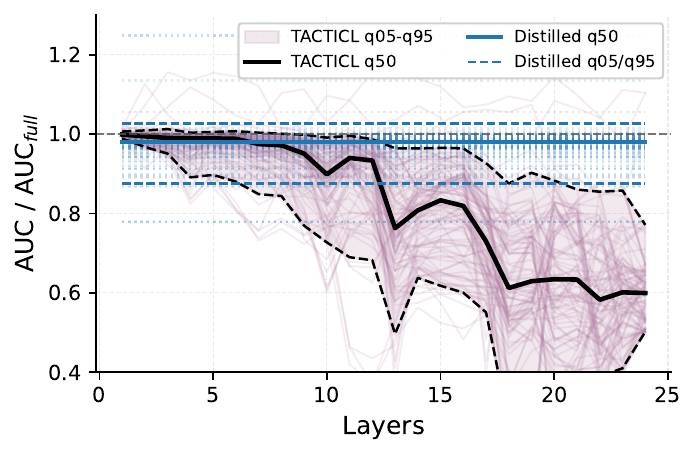}
        \caption{Drop-from-Last}
    \end{subfigure}
    \begin{subfigure}{0.3\linewidth}
        \includegraphics[width=\linewidth]{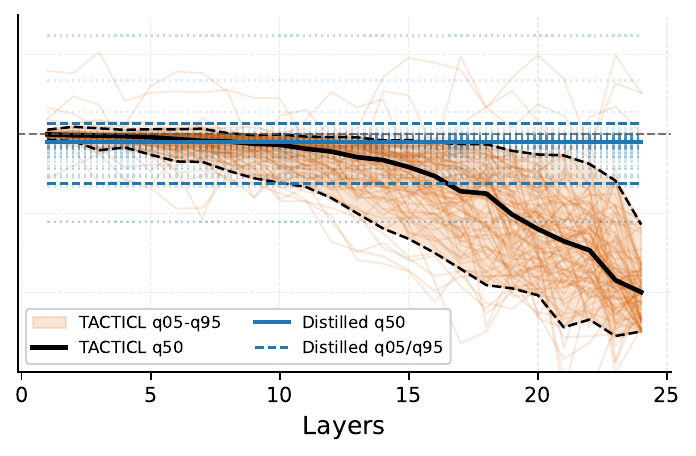}
        \caption{AUC-guided}
    \end{subfigure}
    \begin{subfigure}{0.3\linewidth}
        \includegraphics[width=\linewidth]{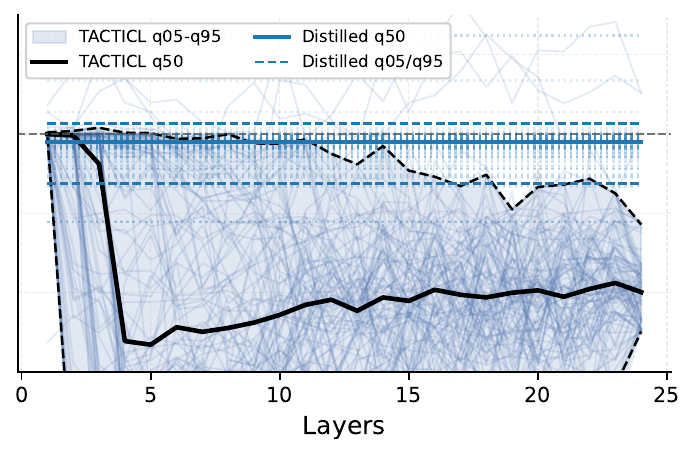}
        \caption{Stability-guided}
        \label{fig:scaling-pfn2.5-cls-del-stab}
    \end{subfigure}
    \caption{\methodname{} deletion applied to \pfntwohalf{}. We report the ratio in performance drop in AUC aggregated across \tabarenac{} datasets for dropping layers across 3 strategies.
    \label{fig:scaling-pfn2.5-cls-del}}
\end{figure}
\begin{figure}[h]
    \centering
    \begin{subfigure}{0.3\linewidth}
        \includegraphics[width=\linewidth]{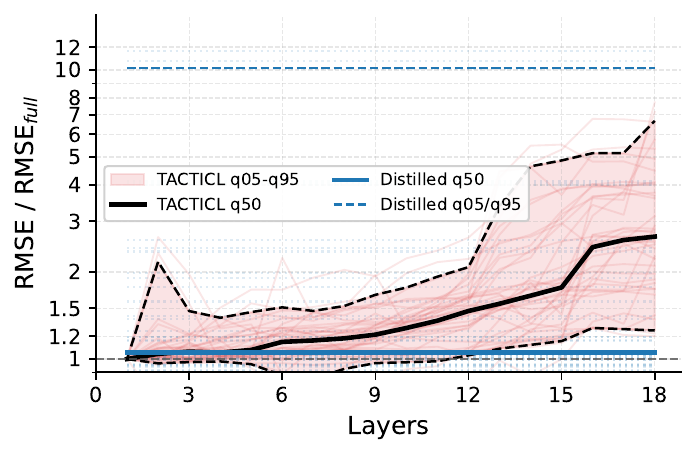}
        \caption{Sub-from-Last}
    \end{subfigure}
    \begin{subfigure}{0.3\linewidth}
        \includegraphics[width=\linewidth]{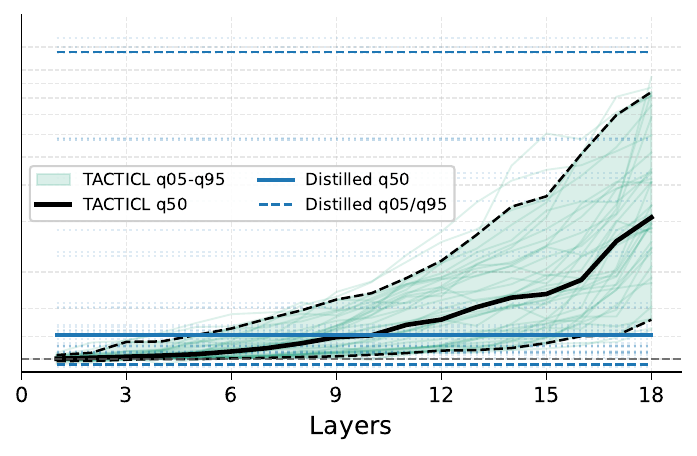}
        \caption{RMSE-guided}
        \label{fig:scaling-pfn2.5-reg-sub-rmse}
    \end{subfigure}
    \begin{subfigure}{0.3\linewidth}
        \includegraphics[width=\linewidth]{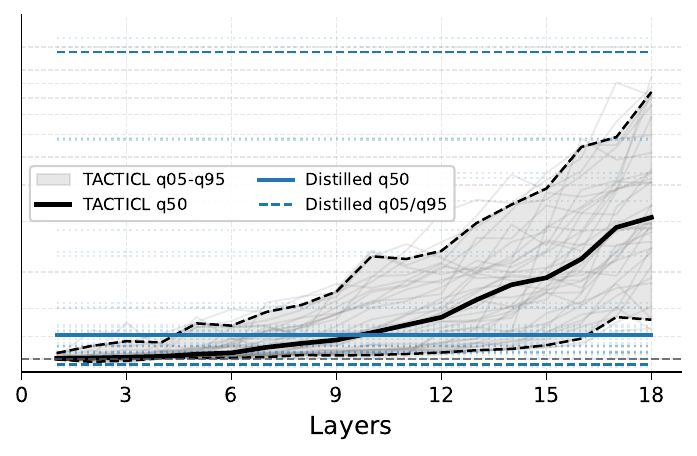}
        \caption{Stability-guided}
    \end{subfigure}
    \caption{\methodname{} substitution applied to \pfntwohalf{}. We report the ratio in performance drop in RMSE aggregated across \tabarenar{} datasets for substituting layers across 3 strategies.
    \label{fig:scaling-pfn2.5-reg-sub}}
\end{figure}
\begin{figure}[h]
    \centering
    \begin{subfigure}{0.3\linewidth}
        \includegraphics[width=\linewidth]{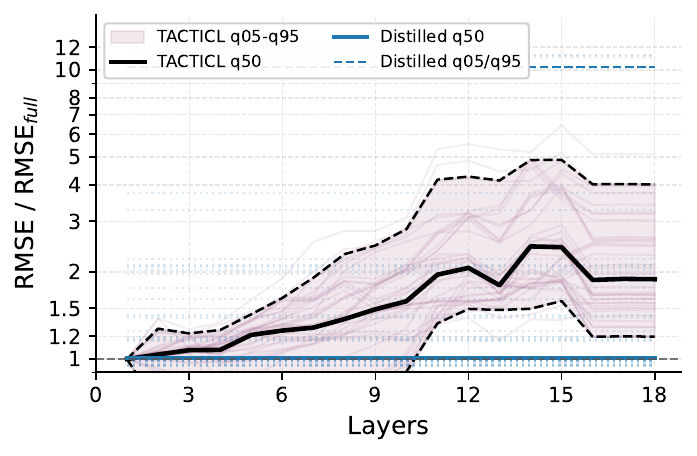}
        \caption{Drop-from-Last}
    \end{subfigure}
    \begin{subfigure}{0.3\linewidth}
        \includegraphics[width=\linewidth]{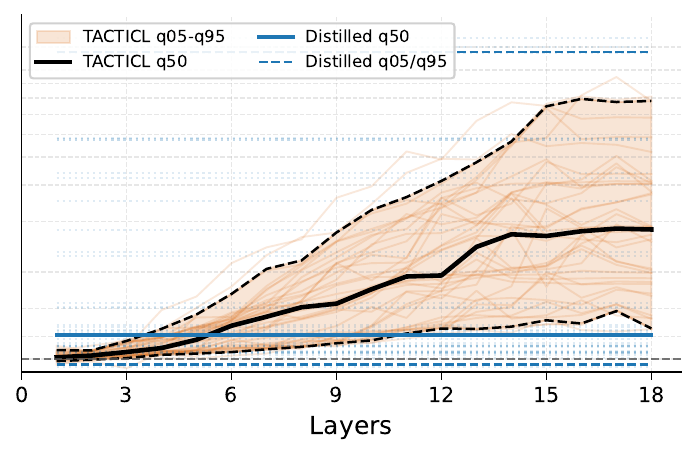}
        \caption{RMSE-guided}
    \end{subfigure}
    \begin{subfigure}{0.3\linewidth}
        \includegraphics[width=\linewidth]{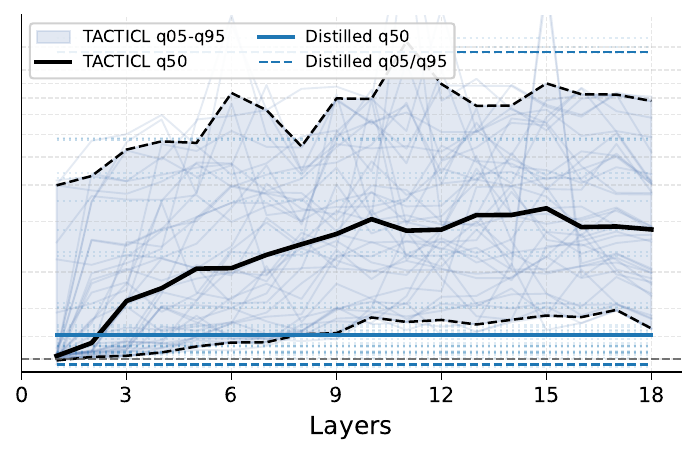}
        \caption{Stability-guided}
    \end{subfigure}
    \caption{\methodname{} deletion applied to \pfntwohalf{}. We report the relative difference in RMSE aggregated across \tabarenar{} datasets for dropping layers across 3 strategies.
    \label{fig:scaling-pfn2.5-reg-del}}
\end{figure}

\begin{figure}[h]
    \centering
    \begin{subfigure}{0.4\linewidth}
        \includegraphics[width=\linewidth]{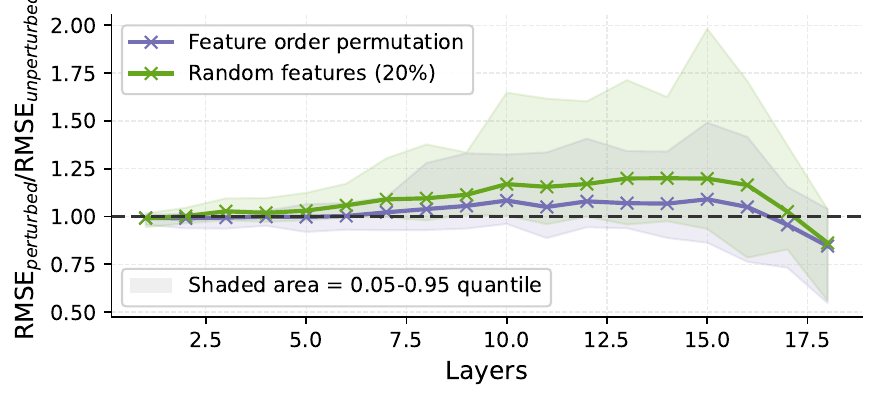}
        \caption{Perturbations evaluation}
        \label{fig:perturbation-reg}
    \end{subfigure}
    \begin{subfigure}{0.4\linewidth}
        \includegraphics[width=\linewidth]{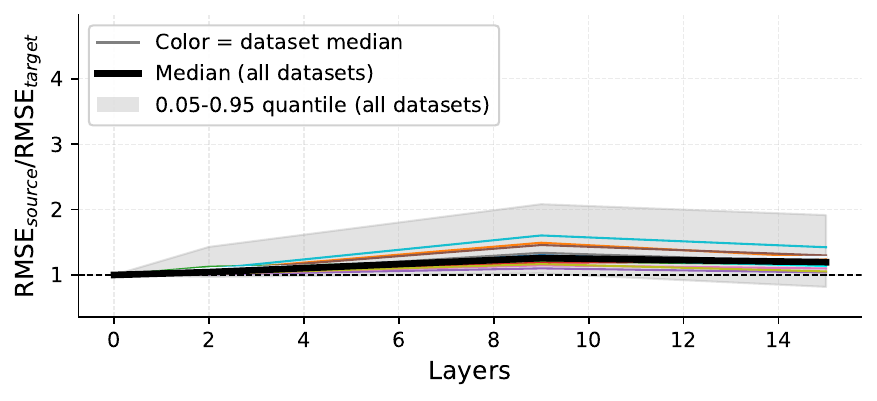}
        \caption{Cross-dataset Evaluation}
        \label{fig:cross-dataset-reg}
    \end{subfigure}
    \caption{Evaluations on perturbed datasets (OOD) (\ref{fig:perturbation-reg}) and cross-dataset generalization (\ref{fig:cross-dataset-reg}) confirm that the compressed model retains its ICL capabilities beyond the training distribution.}
    \label{fig:ood-reg}
\end{figure}


\end{document}